\documentclass[letterpaper, 10 pt, conference]{ieeeconf}  %

\IEEEoverridecommandlockouts                              %

\usepackage{amsmath}
\usepackage{hyperref}
\usepackage{tikz}
\usetikzlibrary{shapes,arrows, calc, arrows.meta, intersections, positioning, patterns, decorations.pathreplacing, backgrounds} %
\usepackage{selectp} %
\usepackage{mathtools}
\usepackage{subcaption}
\usepackage{csquotes}
\usepackage{amsfonts}
\usepackage[noend]{algorithm2e}
\usepackage{calculator}
\usepackage[absolute,overlay]{textpos}
\usepackage{siunitx}
\usepackage{cite}
\usepackage{makecell, multirow}
\usepackage{comment}
\setcellgapes{1.1pt}
\usepackage{booktabs}
\newcommand{\tabitem}{~~\llap{\textbullet}~~}

\newcommand*\rot{\rotatebox{90}}
\usepackage{pifont}
\definecolor{TABLE_GOOD}{RGB}{0,100,0}
\definecolor{TABLE_BAD}{RGB}{214,39,40}
\definecolor{PATH_OKAY}{RGB}{0,0,0}
\definecolor{PATH_COLLISION}{RGB}{185,0,0}
\definecolor{PLOT_RED}{RGB}{214,39,40}
\definecolor{PLOT_RED}{RGB}{214,39,40}
\definecolor{PLOT_BLUE}{RGB}{0,0,150}
\definecolor{LINE_COLOR_RED}{RGB}{214,39,40}
\definecolor{LINE_COLOR_BLUE}{RGB}{0,0,150}
\definecolor{LINE_COLOR_GREEN}{RGB}{44,160,44}
\definecolor{REFERENCE_PATH_GREEN}{RGB}{65,251,61}
\definecolor{REFERENCE_PATH_PURPLE}{RGB}{124,50,135}
\definecolor{OBSERVATION_PATH_CYAN}{RGB}{81,171,184}
\definecolor{GENERATED_PATH_RED}{RGB}{254,24,1}
\newcommand{\yes}{\textcolor{TABLE_GOOD}{\ding{51}}}
\newcommand{\no}{\textcolor{PATH_COLLISION}{\ding{55}}}
\def\linkToCode{www.github.com/translearn/pathTracking}
\def\linkToVideo{https://youtu.be/gCPN8mqPVHg}
\RestyleAlgo{ruled}

\tikzstyle{image_frame} = [rounded corners=1.5pt, inner sep=0.25pt, draw=black]

\title{\LARGE \bf
Learning Time-optimized Path Tracking\\ with or without Sensory Feedback  %
}

\author{Jonas C. Kiemel$^{1}$ and Torsten Kröger %
\thanks{\protect\hypertarget{link:author}{$^{1}$}Institute for Anthropomatics and Robotics – Intelligent Process Automation and Robotics (IAR-IPR), Karlsruhe Institute of Technology (KIT), jonas.kiemel@kit.edu
	 \indent $^{2}$Code: \url{\linkToCode}  \indent $^{3}$Video: \url{\linkToVideo} 
	}%
}

\begin{document}

\maketitle
\thispagestyle{empty}
\pagestyle{empty}

\begin{textblock*}{14.9cm}(3.2cm,0.75cm) 
	{\footnotesize © 2022 IEEE.  Personal use of this material is permitted.  Permission from IEEE must be obtained for all other uses, in any current or future media, including reprinting/republishing this material for advertising or promotional purposes, creating new collective works, for resale or redistribution to servers or lists, or reuse of any copyrighted component of this work in other works.}
\end{textblock*}

\vspace*{-0.6cm}
\begin{abstract}
In this paper, we present a learning-based approach that allows a robot to quickly follow a reference path defined in joint space without exceeding limits on the position, velocity, acceleration and jerk of each robot joint. 
Contrary to offline methods for time-optimal path parameterization, the reference path can be changed during motion execution.
In addition, our approach can utilize sensory feedback, for instance, to follow a reference path with a bipedal robot without losing balance. %
With our method, the robot is controlled by a neural network that is trained via reinforcement learning using data generated by a physics simulator.
From a mathematical perspective, the problem of tracking a reference path in a time-optimized manner is formalized as a Markov decision process. 
Each state includes a fixed number of waypoints specifying the next part of the reference path.
The action space is designed in such a way that all resulting motions comply with the specified kinematic joint limits.
The reward function finally reflects the trade-off between the execution time, the deviation from the desired reference path and optional additional objectives like balancing. %
We evaluate our approach with and without additional objectives and show that time-optimized path tracking can be successfully learned for both industrial and humanoid robots. 
In addition, we demonstrate that networks trained in simulation can be successfully transferred to a \mbox{real robot.} 
A video presentation is available at \normalfont{\url{https://youtu.be/hBukfMs6We8}}.

\end{abstract}

\section{INTRODUCTION}

\begin{figure}[t]
\vspace{-0.4cm}
\def\figureYOffset{-13.95cm}
\def\figureYOffsetBottom{-6.45cm}
\captionsetup[subfigure]{margin=80pt}
\begin{tikzpicture}
                \def\xminPlot{0.151} 
                \def\xmaxPlot{0.949} 
                \SUBTRACT{\xmaxPlot}{\xminPlot}{\xdeltaPlot}
                \DIVIDE{\xdeltaPlot}{3}{\xdeltaPlotNorm}
                \def\yminPlot{0.135}
                \def\ymaxPlot{0.985}
                \def\yminAcc{0.364}
                \def\ymaxAcc{0.535}
                \SUBTRACT{\ymaxPlot}{\yminPlot}{\ydeltaPlot}
                \SUBTRACT{\ymaxAcc}{\yminAcc}{\ydeltaAcc}
                \MULTIPLY{\ydeltaAcc}{4}{\ydeltaGraph}
                \SUBTRACT{\ydeltaPlot}{\ydeltaGraph}{\ydeltaTmp}
                \DIVIDE{\ydeltaTmp}{3}{\ydeltaGap}
                
                 \node[text width=4cm] at (0.0cm, 0cm) (origin){};
                	\node[align=left, scale=0.9] at ($(origin.center)+(-0.15cm, -0.275cm)$) {Without sensory feedback};
                	\node[align=left, scale=0.9] at ($(origin.center)+(4.35cm, -0.275cm)$) {With sensory feedback};
            	\draw [draw=black!50, line width=1.0pt, scale=0.7] ($(origin.center)+(3.075cm, -0.1cm)$) -- + (0.0cm, -13.0cm) node[pos=1, right, yshift=0.01cm, align=left, scale=0.85]{};
    \end{tikzpicture}
    \begin{subfigure}[c]{0.23\textwidth}
        \vspace{\figureYOffset}
	   \vspace{-0.0cm}
	   \begin{tikzpicture}[scale=1.0, every node/.style={scale=1.0}, node distance=2cm]
            \node[image_frame] at (0, 0.0cm) { \includegraphics[trim=800 340 800 220, clip, height=0.92\textwidth]{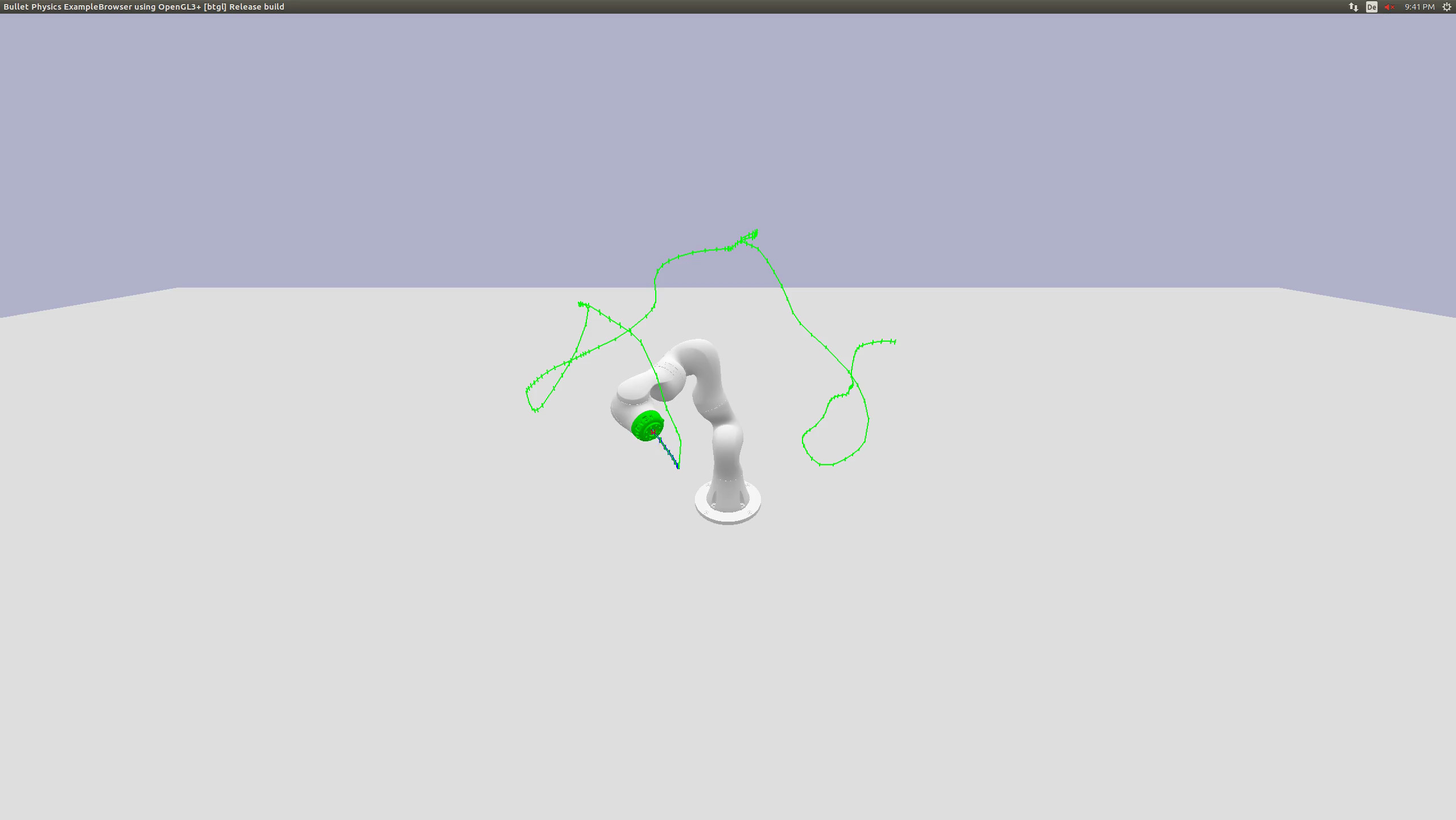}};
       \end{tikzpicture}

	   \vspace{-0.5cm}\hspace*{-2.5cm}\subcaptionbox{KUKA iiwa}[9cm]

	\end{subfigure}
	\hspace{0.0065\textwidth}
	\begin{subfigure}[c]{0.23\textwidth}
	    \vspace{\figureYOffset}
	    \vspace{-0.0cm}
	    \begin{tikzpicture}[scale=1.0, every node/.style={scale=1.0}, node distance=2cm]
            \node[image_frame] at (0, 0.0cm) {\includegraphics[trim=800 340 800 220, clip, height=0.92\textwidth]{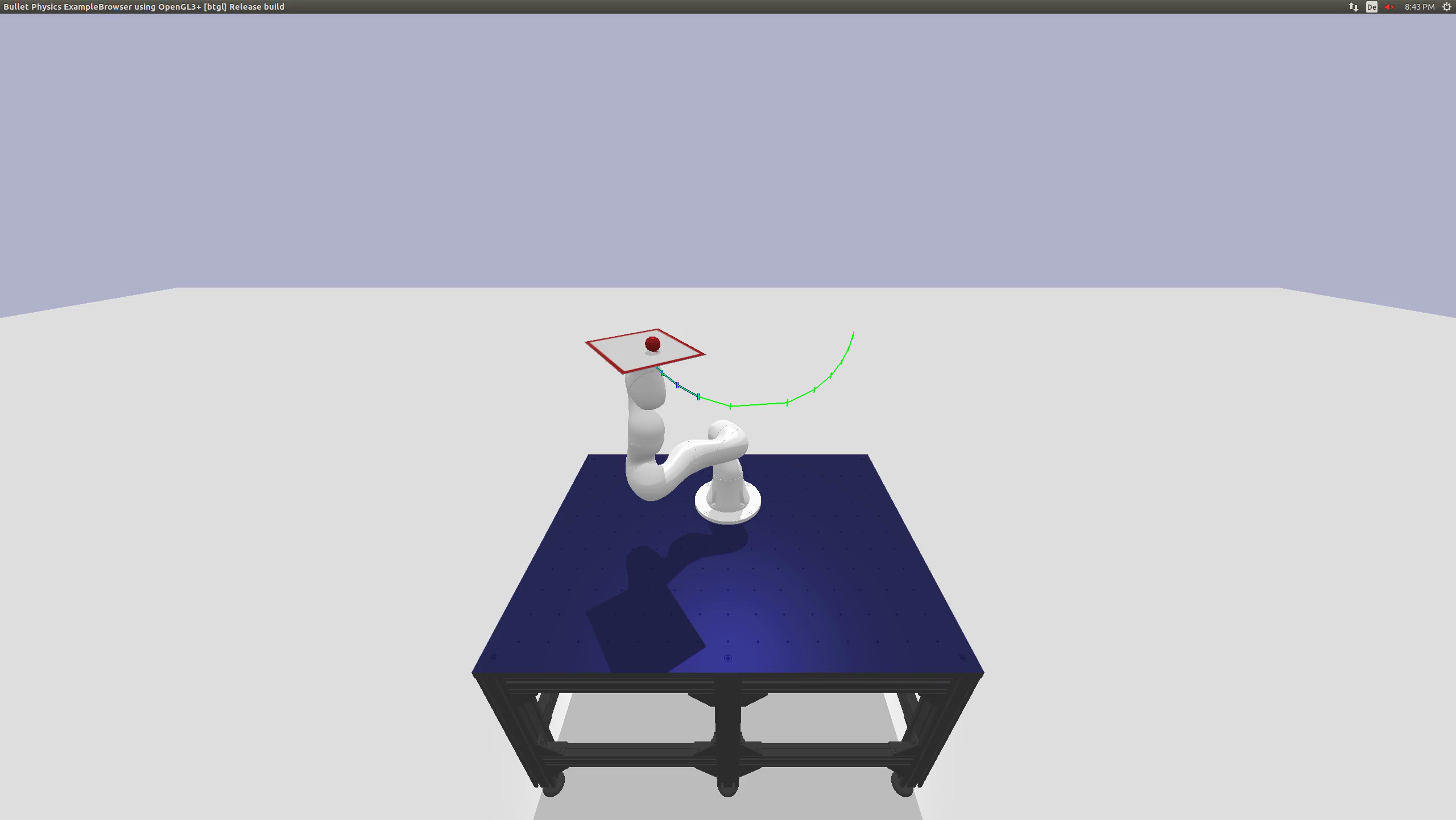}};
       \end{tikzpicture}

		\vspace{-0.5cm}\hspace*{-2.94cm}\subcaptionbox{KUKA with balance board}[10cm]

	\end{subfigure} 
	
	\vspace{0.18cm}
	
    \begin{subfigure}[c]{0.23\textwidth}
       \vspace{\figureYOffsetBottom}
	   \vspace{0.0cm} 
	   \begin{tikzpicture}[scale=1.0, every node/.style={scale=1.0}, node distance=2cm]
            \node[image_frame] at (0, 0.0cm) {\includegraphics[trim=800 450 800 110, clip, height=0.92\textwidth]{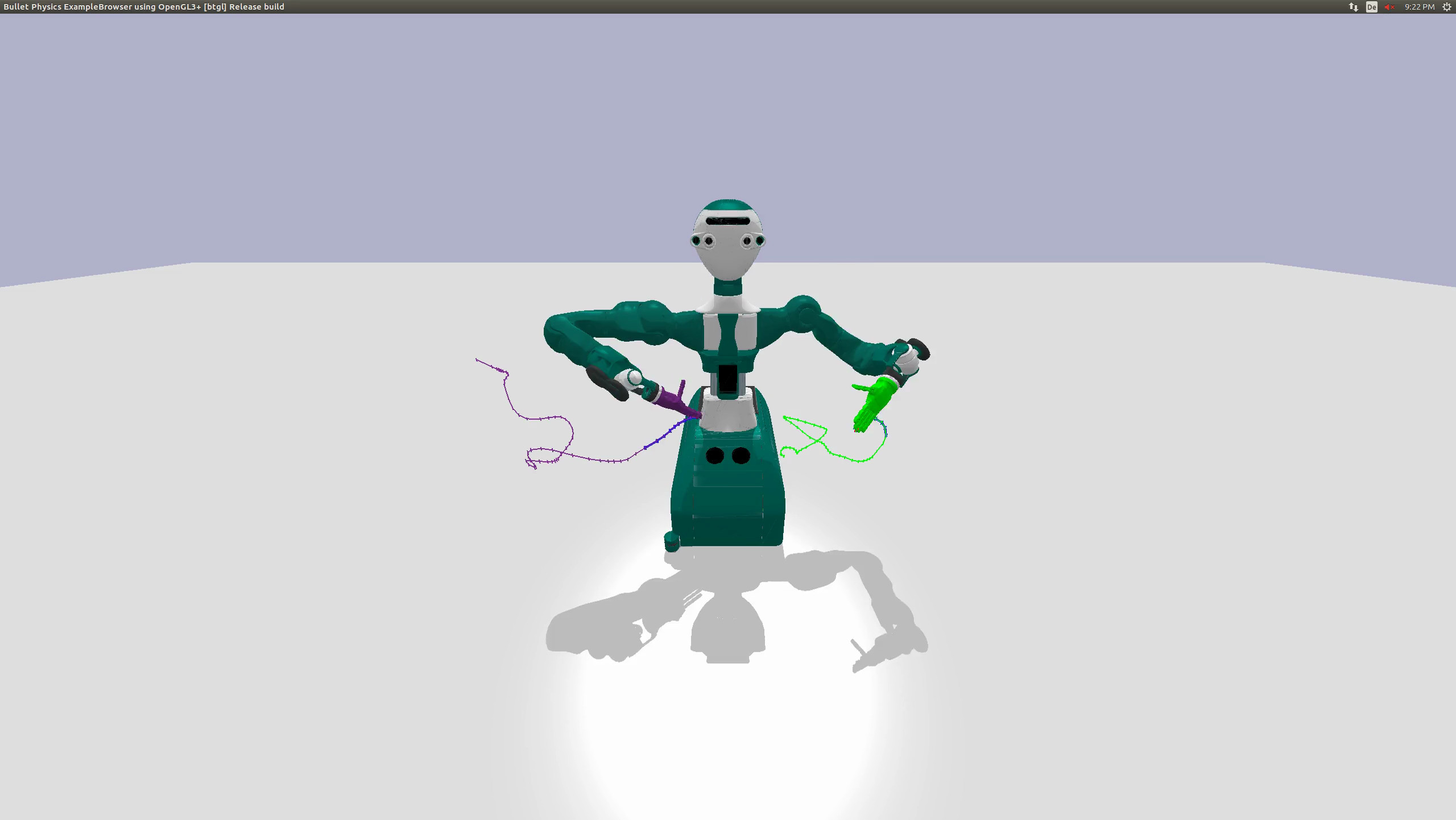}};
       \end{tikzpicture}

	   \vspace{-0.425cm}\hspace*{-2.05cm}\subcaptionbox{ARMAR-6}[8cm]

	\end{subfigure}
	\hspace{0.0065\textwidth}
	\begin{subfigure}[c]{0.23\textwidth}
	    \vspace{\figureYOffsetBottom}
	    \vspace{0.0cm}
	    \begin{tikzpicture}[scale=1.0, every node/.style={scale=1.0}, node distance=2cm]
            \node[image_frame] at (0, 0.0cm) { \includegraphics[trim=790 435 760 80, clip, height=0.92\textwidth]{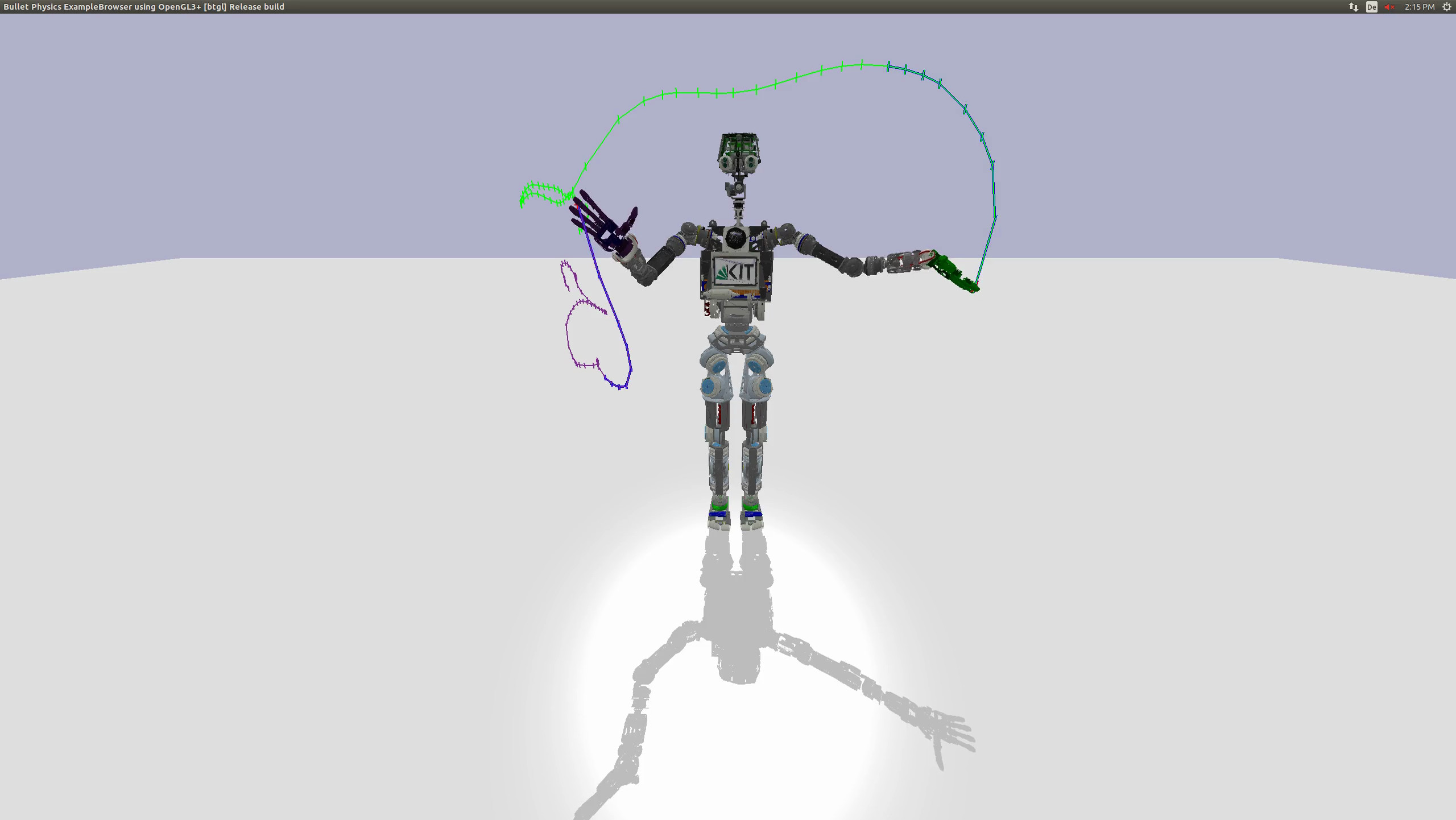}};
        \end{tikzpicture}

		\vspace{-0.425cm}\hspace*{-2.55cm}\subcaptionbox{ARMAR-4}[9cm]

	\end{subfigure} 
	\vspace{-1.2cm}   
	\caption{%
	Our evaluation environments for learning time-optimized path tracking.
	Using sensory feedback, our method can incorporate additional goals such as ball balancing or maintaining balance with the bipedal \mbox{robot ARMAR-4}.
	}%
	\vspace{-0.7cm}   
	\label{fig:header}
	
\end{figure}
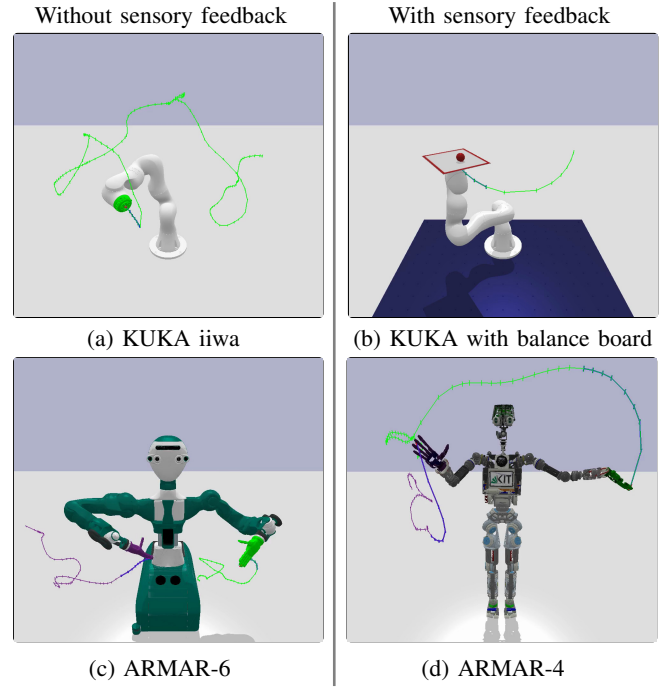

Finding a time-optimal way to follow a reference path while respecting kinematic joint limits is one of the most prevalent tasks in industrial robotics. 
However, just like a race driver needs to know the exact course of the race track to achieve an optimal lap time, time-optimal path parameterization (TOPP) requires the desired reference path to be fully known in advance. 
For that reason, existing offline methods for time-optimal path parameterization are not applicable to reactive online scenarios in which the reference path needs to be adjustable during motion execution. 
The same is true if sensory feedback is needed to successfully perform the path tracking task.
As shown in Fig.~\ref{fig:header}, this applies, for example, to the bipedal humanoid robot ARMAR-4, which can lose its balance when its arms are moved or to an industrial robot that balances a ball on a plate while following a reference path.
In this work, we address the problem of time-optimized path tracking for online scenarios like those shown in Fig. \ref{fig:header} by learning a well-performing trade-off between the execution speed, the deviation from the reference path and additional objectives like ball balancing via model-free reinforcement learning (RL). 
For this purpose, a neural network is trained in a simulation environment using thousands of different reference paths. 
At each time step, only the directly following part of the reference path is made available to the neural network as an input signal.
Once the training process is finished, the neural network can generate optimized trajectories even for paths not included in the training set. Our action space ensures that all generated trajectories comply with predefined kinematic joint limits. 
Additional optimization objectives like ball balancing can be easily added to the reward function as they do not have to be differentiable with respect to the selected action. \\
Our main contributions can be summarized as follows: 
\begin{itemize}
\item We introduce an online method to track reference paths based on reinforcement learning that ensures compliance with kinematic joint limits at all times. 
\item We show that our approach can incorporate sensory feedback to account for additional objectives such as maintaining the balance of a bipedal humanoid robot. 
\item We evaluate our method using robots with up to 30 degrees of freedom and demonstrate successful sim-2-real transfer for a time-optimized ball-on-plate task. %
\end{itemize}
Our source code is publicly available.\href{https://\linkToCode}{$^2$}

\begin{figure*}[t]
    \input{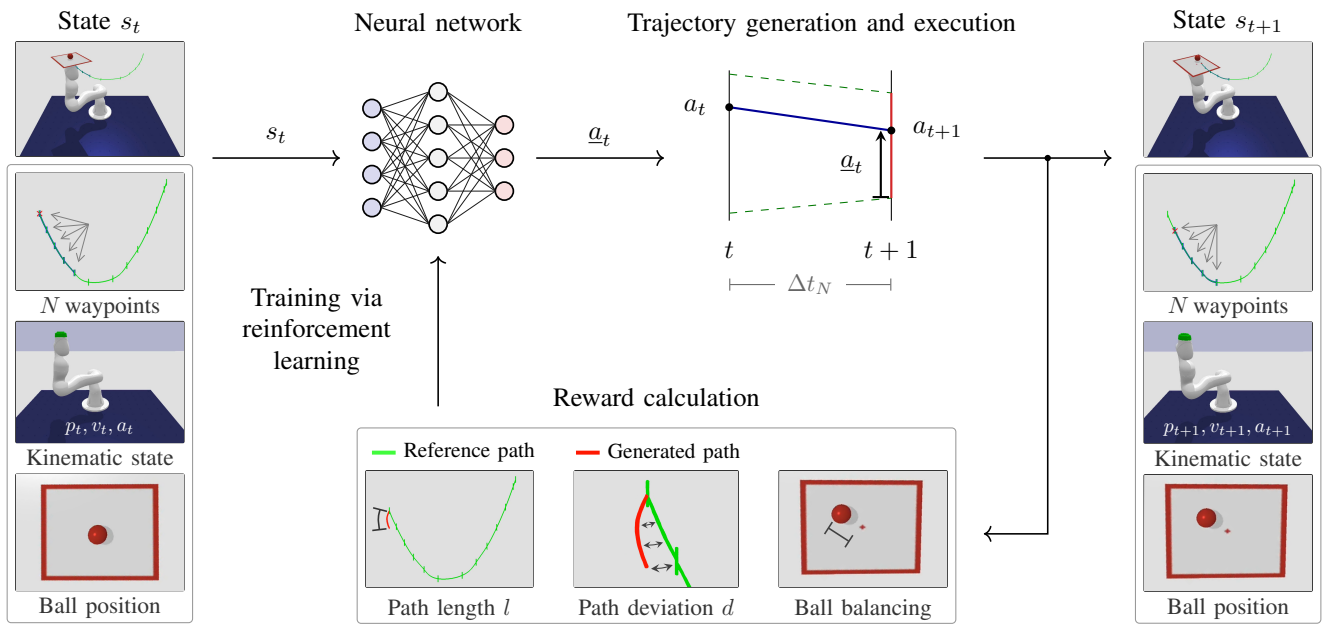}
	\caption{%
	The principle of online trajectory generation with our approach illustrated for a single time step $t$ of a \mbox{ball-on-plate task.}} 
	\label{fig:basic_principle}
	\vspace{-0.5cm}
\end{figure*}

\section{Related work}

\subsection{Time-optimal path parameterization (TOPP)}
The problem of finding a time-optimal path parameterization subject to kinematic joint constraints has been studied for decades with methods proposed based on dynamic programming \cite{shin85Time}, numerical integration \cite{bobrow1985time}, convex optimization \cite{verscheure2009time} and reachability analysis \cite{pham2018Toppra}.  
Widely used implementations that can consider velocity and acceleration constraints include \cite{kunz2012time} for paths consisting of line segments and circular blends and \cite{pham2018Toppra} for paths defined as cubic splines. 
While our approach additionally supports jerk limits, TOPP subject to jerk constraints is still an open research problem~\cite{pham2017structure}. 
Methods related to TOPP have also been used for ball balancing \cite{kiemel2020truerma} or to keep bipedal humanoid robots in balance e.g. by defining multi-contact friction constraints \cite{hauser2014fast, pham2018Toppra} or by imposing constraints on the zero moment point (ZMP) \cite{pham2012time}.
However, as time-optimal path parameterization requires the desired path to be known in advance, all calculations have to be performed offline. As a consequence, it is not possible to consider sensory feedback to compensate for external disturbances or model errors. The latter is particularly problematic as existing methods often require simplified contact models. 

\subsection{Online trajectory generation (OTG)}
Time-optimal point-to-point motions subject to kinematic joint constraints can be computed online, for instance using the Reflexxes motion library \cite{kroger2011opening}. This is done by specifying a desired kinematic target state at each control time step. In contrast to our work, this method does not allow to specify a desired reference path. 
An online method for computing a feasible and jerk-limited path-accurate robot trajectory is presented in \cite{lange2015path}.
However, unlike our work, the method does not put focus on generating time-optimized trajectories. %
Another line of research aims at finding optimized robot trajectories based on model-predictive control (MPC) \cite{faulwasser2016implementation, carron2019data, nubert2020safe}.
Compared to MPC, model-free RL offers the advantage that no differentiable dynamics model is required and that the reward function does not have to be differentiable with respect to the selected action.
In addition, generating optimized motions with neural networks trained via RL is very fast, which allows real-time execution even for systems with many degrees of freedom. 
As a consequence, model-free RL can be used for a wide range of applications and has attracted increasing attention from the research community over the past few years. %
Exemplary learning tasks presented in the past include in-hand manipulation~\cite{andrychowicz2020learning}, locomotion with bipedal robots \cite{siekmann2021sim} or collision-free target point reaching~\cite{kiemel2021collision}, to name just a few.
Model-free RL has also been used to adjust reference trajectories in order to meet additional objectives \cite{peng2018deepmimic, kiemel2020trueadapt}. Compared to these approaches, we consider reference paths instead of reference trajectories, meaning that the velocity profile of the resulting trajectory is part of the optimization rather than being specified in advance. 
Considering that additional objectives like balancing become more difficult the faster the reference path is traversed, optimizing the traversing speed during the training process is an important factor for learning well-performing trajectories. 

\label{sec:citations}
\section{Approach}
\subsection{Overview}
Fig.~\ref{fig:basic_principle} illustrates the basic principle of our method based on a ball-on-plate task with sensory feedback. 
The goal of the task is to quickly follow a reference path (shown in green) while avoiding a ball to fall from a balance board attached to the last link of the robot.
The position of the ball on the board is measured and provided as sensory feedback. 
Using the mathematical framework of a Markov decision process, we formalize the online generation of optimized robot trajectories as a discrete-time control problem with a constant time~$\Delta t_N$ between decision steps. %
At each decision step~$t$, a state description $s_t$ containing information on the reference path, the kinematic state of the robot and the ball position is given as input to a neural network. 
The neural network outputs an action~$\underline{a}_{t}$ that is used to compute a continuous trajectory from the current time step~$t$ to the subsequent time step $t+1$. 
During training, the trajectory is executed in a physics simulator and a scalar reward is computed to evaluate the performance of the generated action~$\underline{a}_{t}$. 
As part of the reward computation, the path resulting from the trajectory execution (shown in red) is assessed based on its length and its deviation to the reference path. 
In addition, the balancing performance is rated based on the distance between the ball and the center of the board.
Using model-free RL, the neural network is trained to output actions that maximize the sum of rewards received over time.
This way, the network learns to generate robot trajectories that are optimized with respect to the execution time, the path deviation and the balancing performance.  
The following subsection provides further details on the Markov decision process.   
Subsequently, we explain how the reference paths used to train the neural network are generated. 
\begin{figure}[b]
	 \centering
	 \vspace{0.0cm}
	 \begin{tikzpicture}[auto, node distance=5cm,>=latex', scale=1.0,  every node/.style={scale=1.0}]
	    \def\textScale{0.9}
	    \node[image_frame](observation_spline) {\includegraphics[trim=500 440 550 1300, clip,   width=0.47\textwidth]{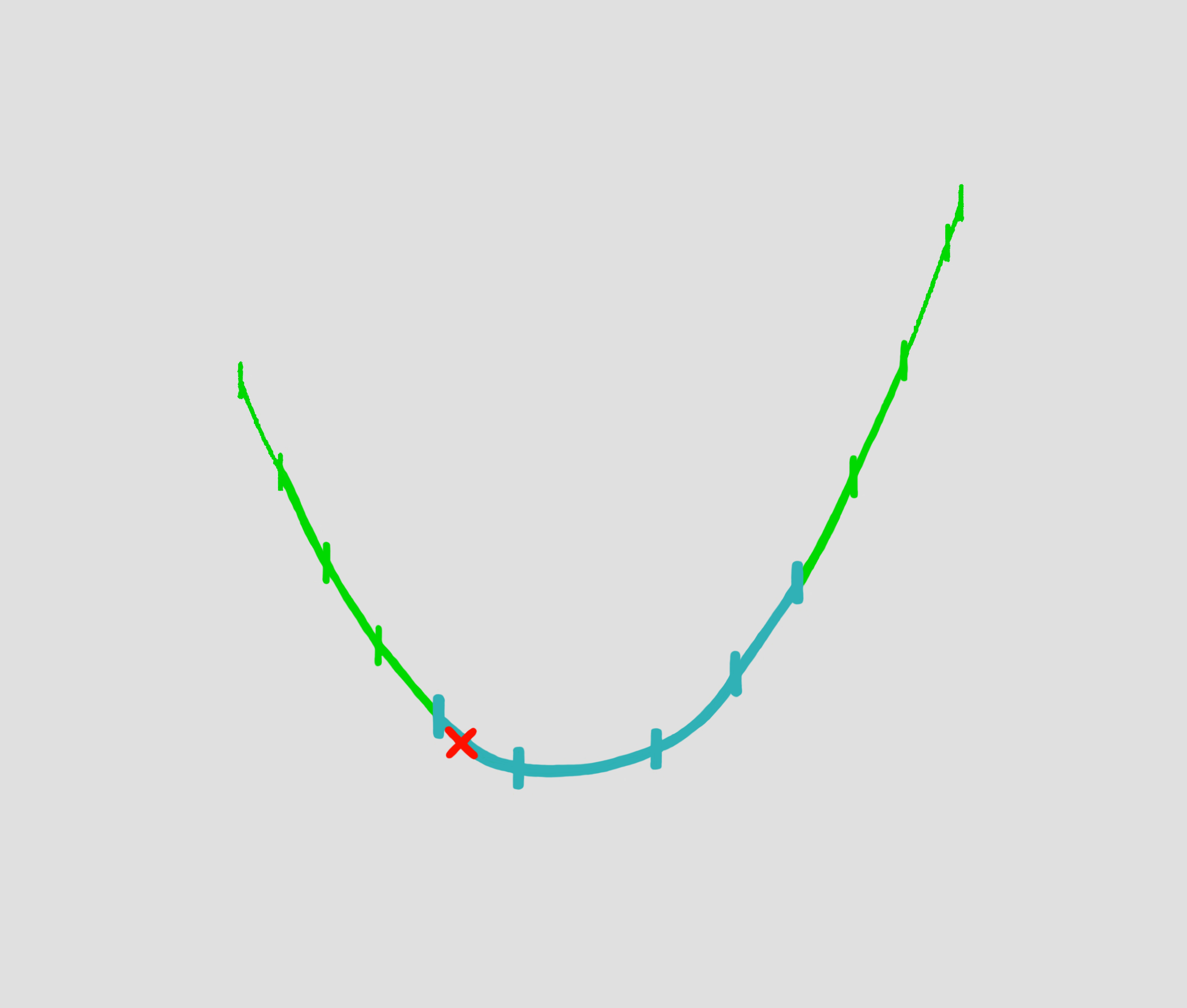}};
	    \node[minimum height=0cm, minimum width=0.0cm](annotation_center) at ($(observation_spline.center) + (0.28cm, -0.15cm)$){};
        \draw [|-|, draw=black!80, thick] plot [smooth, tension=0.5] coordinates { ($(annotation_center.center) + (-1.7cm, 0.125cm)$) ($(annotation_center.center) + (-1.55cm, -0.02cm)$) ($(annotation_center.center) + (-1.4cm, -0.14cm)$)};
        \draw [|-|, draw=black!80, thick] plot [smooth, tension=0.5] coordinates { ($(annotation_center.center) + (-1.85cm, -0.9cm)$) ($(annotation_center.center) + (-0.93cm, -1.35cm)$) ($(annotation_center.center) + (0.6cm, -1.15cm)$)
        ($(annotation_center.center) + (1.5cm, -0.4cm)$)
        ($(annotation_center.center) + (2.4cm, 0.7cm)$)};
        \node[minimum height=0cm, minimum width=0.0cm, align=center, scale=\textScale] at ($(annotation_center.center) + (2.8cm, -0.5cm)$){Path length \\$l_{\textrm{State}}$};
        \node[minimum height=0cm, minimum width=0.0cm, align=center, scale=\textScale] at ($(annotation_center.center) + (-3.25cm, -0.5cm)$){Reference \\ path}; %
        \draw [-stealth, draw=black!80, thick] ($(annotation_center.center) + (-3.2cm, 0.0cm)$) -- ($(annotation_center.center) + (-2.7cm, 0.5cm)$);
         \draw [-stealth, draw=black!80, thick] ($(annotation_center.center) + (-1.85cm, 0.975cm)$) -- ($(annotation_center.center) + (-1.85cm, 0.1cm)$) node[pos=0.0, above, minimum height=0cm, minimum width=0.0cm, xshift=0.05cm, scale=\textScale]{Knot $1$};
         \draw [-stealth, draw=black!80, thick] ($(annotation_center.center) + (1.4cm, 1.175cm)$) -- ($(annotation_center.center) + (1.85cm, 1.175cm)$) node[pos=0.0, left, minimum height=0cm, minimum width=0.0cm, scale=\textScale]{Knot $N$};
         \draw [-stealth, draw=black!80, thick, scale=\textScale] ($(annotation_center.center) + (-1.15cm, 0.05cm)$) -- ($(annotation_center.center) + (-1.5cm, -0.33cm)$) node[pos=0.0, right=-0.0cm, minimum height=0cm, minimum width=0.0cm, align=center, yshift=0.3cm, scale=\textScale]{Current \\path position};
    \end{tikzpicture} 
    \caption{The figure illustrates how the following part of the reference path is included in the state using $N=5$ knots.}
	\label{fig:observation_spline}
\end{figure}
\subsection{Details on the Markov decision process}
The problem of generating optimized robot trajectories is formalized as a Markov decision process $(\mathcal{S}, \mathcal{A}, P_{\underline{a}}, R_{\underline{a}})$, with $\mathcal{S}$ being the state space,  $\mathcal{A}$ being the action space,  $P_{\underline{a}}$~representing (unknown) transition probabilities and $R_{\underline{a}}$~being the reward resulting from action $\underline{a}$. In the following, details on the state space, the action space, the reward calculation and the termination of episodes are provided.  
\subsubsection{Composition of states ${s} \in \mathcal{S}$}

A state ${s}$ contains information on the following part of the reference path and on the kinematic state of the robot.
For tasks that require sensory feedback, the sensor signals are included in the state as well. 
In our work, the reference paths are described as cubic splines. 
A spline is a mathematical representation that can be used to define a piecewise polynomial path between specified waypoints, called knots. 
To provide the neural network with information about the reference path, $N$ of these knots are included in the state.
Note that the reference path is defined in joint space, meaning that each knot is a vector of joint positions. 
For illustration purposes, however, we apply forward kinematics to visualize the reference path in Cartesian space. 
Fig. \ref{fig:observation_spline} illustrates how the knots are selected. 
The red cross indicates the current position on the reference path.
At the beginning of an episode, this position is set to the start of the reference path. 
During an episode, each action causes the robot to move along a certain path. %
After each action, the current position on the reference path is shifted forward according to the length of the path generated by the action.
As shown in Fig. \ref{fig:observation_spline}, the state contains the knot preceding the current path position and $N\!-\!1$ following knots. 
In addition, the path length labeled $l_{\textrm{State}}$ and the path length between the first knot and the current path position are included in the state. 
If the remaining part of the reference path consists of less than $N\!-\!1$ knots, the last knot is inserted into the state multiple times.
In this case, the decreasing path length $l_{\textrm{State}}$ indicates to the network that the robot must be slowed down.
\begin{figure}[t]
\centering
\captionsetup[subfigure]{margin=0pt}
    \vspace{0.2cm}
        \begin{tikzpicture}
                \def\xminPlot{0.151} 
                \def\xmaxPlot{0.949} 
                \SUBTRACT{\xmaxPlot}{\xminPlot}{\xdeltaPlot}
                \DIVIDE{\xdeltaPlot}{3}{\xdeltaPlotNorm}
                \def\yminPlot{0.135}
                \def\ymaxPlot{0.985}
                \def\yminAcc{0.364}
                \def\ymaxAcc{0.535}
                \SUBTRACT{\ymaxPlot}{\yminPlot}{\ydeltaPlot}
                \SUBTRACT{\ymaxAcc}{\yminAcc}{\ydeltaAcc}
                \MULTIPLY{\ydeltaAcc}{4}{\ydeltaGraph}
                \SUBTRACT{\ydeltaPlot}{\ydeltaGraph}{\ydeltaTmp}
                \DIVIDE{\ydeltaTmp}{3}{\ydeltaGap}
                \hspace{0.23cm}
                 \node[text width=4cm] at (0.2cm, 4.35cm) (origin){};
                	\draw [draw=REFERENCE_PATH_GREEN, line width=1.5pt, scale=0.7] ($(origin.center)+(-9.3cm, -0.0cm)$) -- + (0.45cm, 0cm) node[pos=1, right, yshift=0.01cm, align=left, scale=0.85]{Reference path};
                	\draw [draw=OBSERVATION_PATH_CYAN, line width=1.5pt, scale=0.7] ($(origin.center)+(-5.72cm, -0.0cm)$) -- + (0.45cm, 0cm) node[pos=1, right, yshift=0.01cm, align=left, scale=0.85]{Waypoints in state};
            	    \draw [draw=GENERATED_PATH_RED, line width=1.5pt, scale=0.7] ($(origin.center)+(-1.5cm, 0.0cm)$) -- + (0.45cm, 0cm) node[pos=1, right, yshift=0.00cm, align=left, scale=0.85]{Generated path};
    \end{tikzpicture}
    \begin{subfigure}[c]{0.23\textwidth}
	   \vspace{-0.0cm}
	   \begin{tikzpicture}[scale=1.0, every node/.style={scale=1.0}, node distance=2cm]
            \node[image_frame] at (0, 0.0cm) {\includegraphics[trim=450 300 1400 475, clip, height=0.85\textwidth]{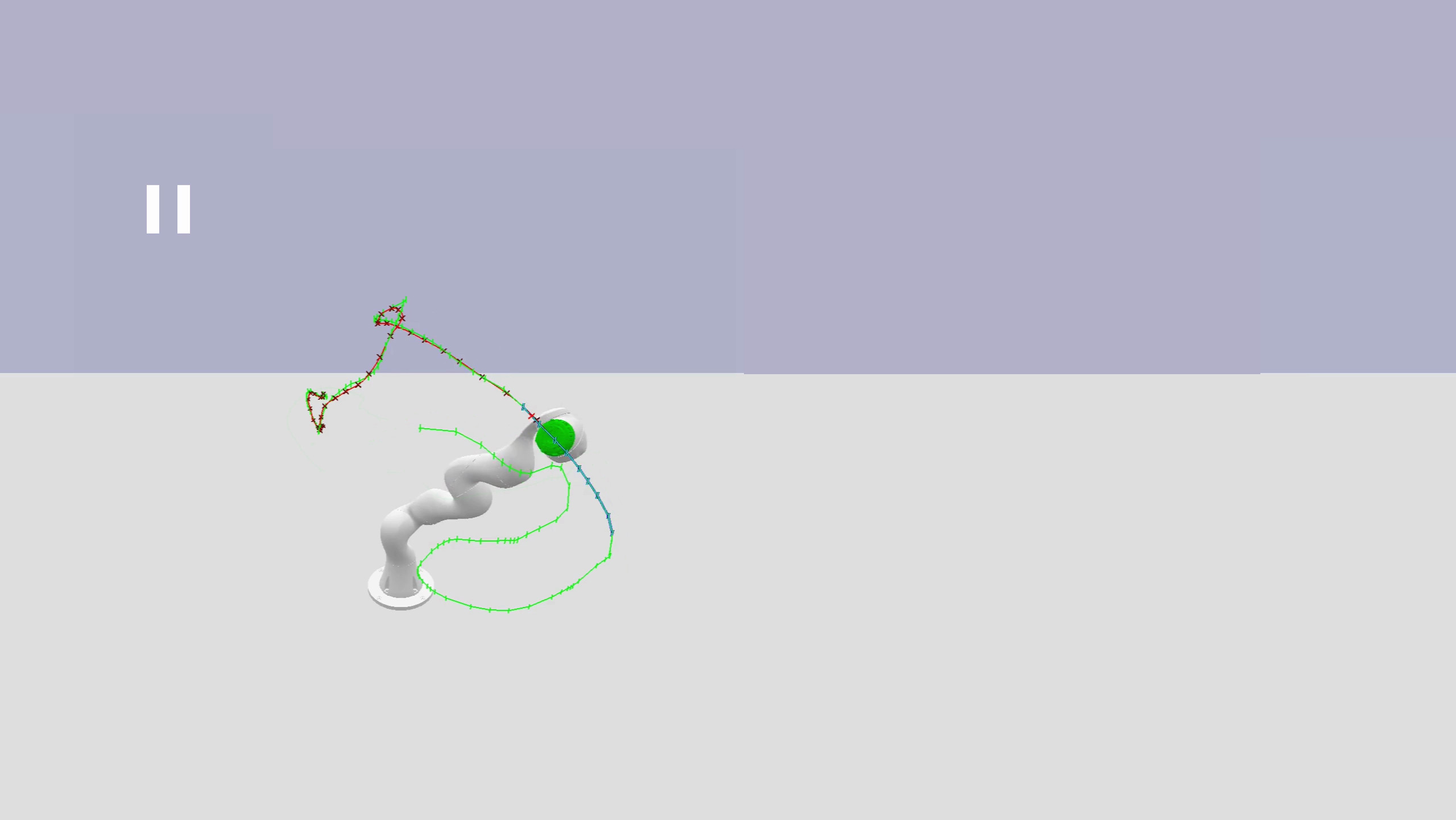}};
       \end{tikzpicture}

	   \vspace{-0.40cm}\hspace*{-0.7cm}\subcaptionbox{Initial reference path}[5cm]

	\end{subfigure}
	\begin{subfigure}[c]{0.23\textwidth}
	    \vspace{-0.0cm}
	    \begin{tikzpicture}[scale=1.0, every node/.style={scale=1.0}, node distance=2cm]
            \node[image_frame] at (0, 0.0cm) { \includegraphics[trim=450 300 1400 475, clip, height=0.85\textwidth]{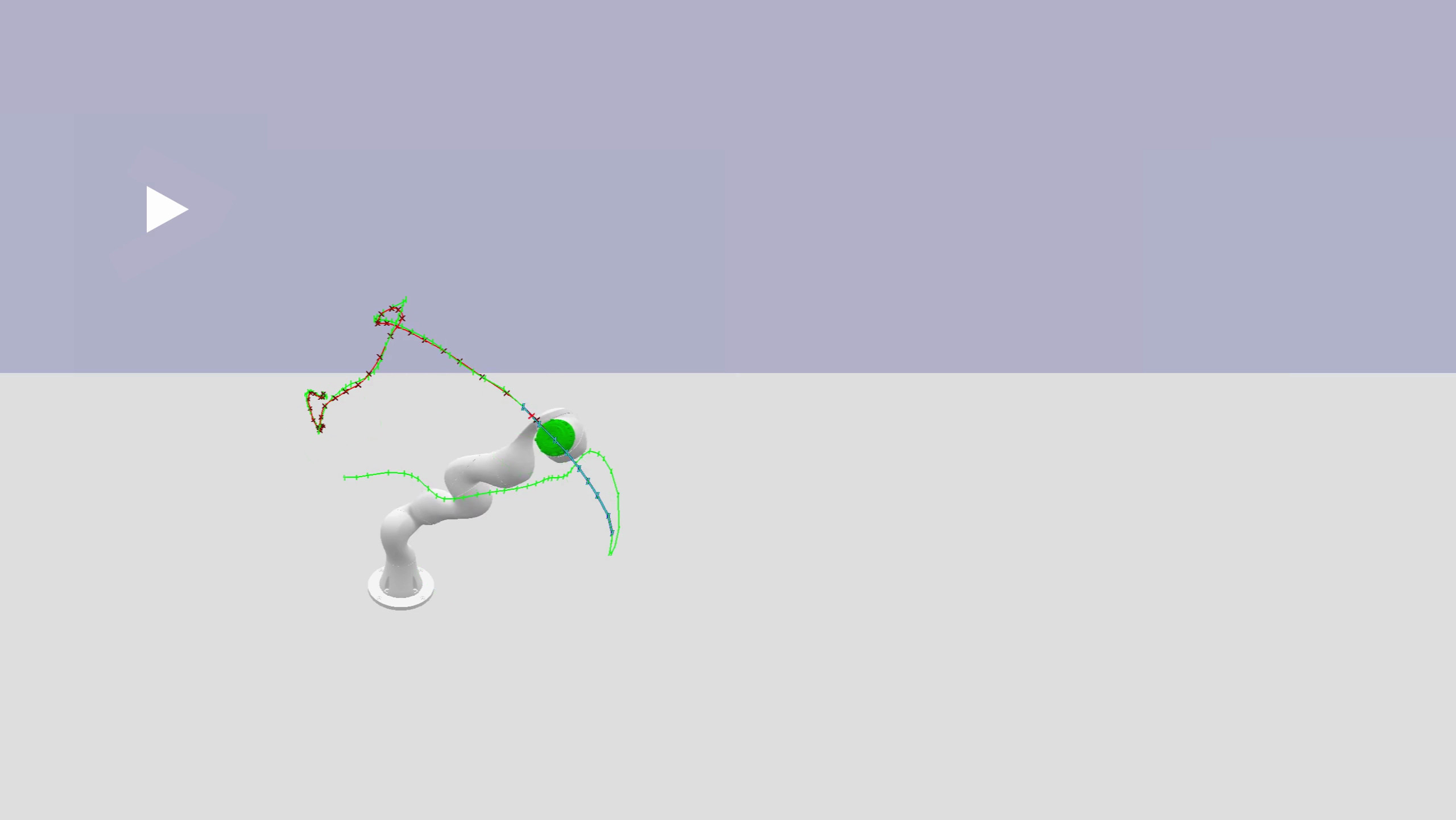}};
        \end{tikzpicture}

		\vspace{-0.4cm}\hspace*{-0.645cm}\subcaptionbox{Adjusted reference path}[5cm]

	\end{subfigure}

	\caption{Adjusting the reference path during motion execution.}
	\label{fig:spline_change}
	\vspace{-0.38cm}
\end{figure}
As illustrated in Fig. \ref{fig:spline_change}, the reference path can be adjusted during motion execution. 
We note that the reference path can be traversed faster if more knots are included in the state.
In return, however, a larger part of the reference path needs to be known in advance.

The kinematic state of the robot is described by the position~$p$, velocity~$v$ and acceleration~$a$ of each robot joint. 

As sensory feedback, the position of the ball on the plate is included in the state of the ball-on-plate task.
To maintain balance with the humanoid robot \mbox{ARMAR-4}, the position and the rotation of the robot's pelvis relative to its initial upright pose are included in the state. While we extract the data from simulation, it could also be provided by an inertia measurement unit (IMU). 
\vspace{0.15cm}
\subsubsection{Trajectory generation based on actions $\underline{a} \in \mathcal{A}$}
\hfill\\
Each action $\underline{a}$ defines a trajectory with a duration of $\Delta t_N$.  
To avoid damage to the robot joints, the following kinematic constraints must be satisfied for each joint at all times:
\begin{alignat}{3}
p_{min} &{}\le{}& \theta &{}\le{}& p_{max}  \label{eq:constraint_p} \\ 
v_{min} &{}\le{}& \dot{\theta} &{}\le{}& v_{max}  \label{eq:constraint_v}\\
a_{min} &{}\le{}& \ddot{\theta}&{}\le{}& a_{max}  \label{eq:constraint_a}  \\
j_{min} &{}\le{}& \dddot{\theta} &{}\le{}& j_{max},  \label{eq:constraint_j}
\end{alignat}
where $\theta$ is the joint position and $p$, $v$, $a$ and $j$ stand for position, velocity, acceleration and jerk, respectively.
\begin{figure}[t]
	\vspace{0.1cm}
	 \begin{tikzpicture}[auto, node distance=5cm,>=latex', scale=0.7,  every node/.style={scale=0.85}]
        \node[text width=4cm] at (0, 0) (origin){};
    	\draw [draw=LINE_COLOR_BLUE, line width=1.5pt] ($(origin.center)+(-3.8cm, -0.0cm)$) -- + (0.45cm, 0cm) node[pos=1, right, yshift=-0.01cm, align=left]{\small{Generated trajectory}};
	    \draw [solid, draw=LINE_COLOR_RED, line width=1.5pt] ($(origin.center)+(0.4cm, 0.0cm)$) -- + (0.45cm, 0cm) node[pos=1, right, yshift=-0.01cm, align=left]{\small{Kinematically feasible acceleration setpoints}};
    \end{tikzpicture} 
    
	\hspace{-5.5cm}
	\resizebox{0.8\textwidth}{!}{
    	\def\ymax{2.5}
	\def\ymin{-2.4}
	\def\xdelta{2.955}
	\def\xmax{3*\xdelta + 0.5}
	\def\amin{-2.2}
	\def\amax{2.2}
	
	\def\aonemaxA{1.3}
	\def\aonemaxB{1.1}
	\def\jmin{-2.6} 
	\def\aonemaxC{\amin-\jmin}
	\def\astarB{-2.0}
	\def\astarC{-0}
	\def\jmax{0.65}
	\def\njmaxFractionC{0.75}
	\def\vstarminusoneA{1.2}
	\def\vstarminusoneB{1.0}
	\def\vstarminusoneC{0.8}
	\def\vstarA{-1.2}
	\def\vstarB{-0.4}
	\def\vstarC{0.0}
	\def\tfive{(\x - 5*\xdelta) / \xdelta}
	
	\def\jmaxVis{\jmax*3}
	\def\astarBVis{\astarB*1}
	\def\astarCVis{\astarC*1}
	
	\def\cfiveB{(\vstarminusoneB - \vstarB) / (0.5 * \jmaxVis - \astarBVis)}
	
	\def\dfiveB{\vstarB * (1-\cfiveB)}
	
	\def\cfiveC{(\vstarminusoneC - \vstarC) / (0.5 * \jmaxVis - \astarCVis)}
	\def\dfiveC{\vstarC * (1-\cfiveC)}
	
	\def\steps{3}
	\def\showLimits{1}
	\def\violation{0}
	\def\showRedLine{1}
	\def\showAcc{1}
	
	\def\redLineLinestyle{solid}
    \def\limitsLinestyle{dashed}
	
	\def\azeroMax{1.6}
	\def\aoneMax{1.1}
	\def\atwoMax{0.9}
	\def\athreeMax{1.2}
	\def\afourMax{1.6}
	
	\def\azeroMin{-1.4}
	\def\aoneMin{-0.9}
	\def\atwoMin{-1.3}
	\def\athreeMin{-1.5}
	\def\afourMin{\amin}
	
	\def\azero{0.6}
	
	\def\actionColor{black!75}
	
	\def\aoneViolation{1.5}
	\ifnum \violation = 1
		\def\aone{\aoneViolation}
	\else
		\def\aone{0.5 * \aoneMin + 0.5 * \aoneMax}
	\fi
	
	\def\atwo{0.75 * \atwoMin + 0.25 * \atwoMax}
	\def\athree{1.0 * \athreeMax + 0.0 * \athreeMin}
	\def\afour{1.2}

	\definecolor{POS_LIM_A}{RGB}{0,120,0}%
	\definecolor{POS_LIM_B}{RGB}{0,0,150} %
	\definecolor{POS_LIM_C}{RGB}{0,120,0}%
	\definecolor{POS_LIM_D}{RGB}{214,39,40}
	
	\begin{tikzpicture}[scale=1, every node/.style={scale=1.1}]
	\draw [<-,thick] (0,\ymax) node (yaxis) [above] {$a$} -- (0,\ymin) node[below=0.1cm, name=nodet0] {$t$};
	
	\draw[dashed] (\xdelta, \ymax)  -- (\xdelta, \ymin) node[below=0.1cm] {$t+1$};
	\draw[dashed] (2*\xdelta, \ymax)  -- (2*\xdelta, \ymin) node[below=0.1cm] {$t+2$};
	\draw[dashed] (3*\xdelta, \ymax)  -- (3*\xdelta, \ymin) node[below=0.1cm] {$t+3$};
	
	\draw[dashed] (\xdelta, \ymax - 0.1)  -- + (\xdelta, 0) node[pos=0.5, fill=white, scale=0.9] {\textcolor{black!70}{$\Delta t_N$}};

	\ifnum \showLimits > 0
		\draw[\limitsLinestyle, color=POS_LIM_A, very thick] (0,\azeroMax) node[left=-0.0275cm, color=black] {$\boldsymbol{a_{t_{max}}}$} -- (\xdelta,\aoneMax); 
		\ifnum \steps > 1
			\draw[\limitsLinestyle, color=POS_LIM_A, very thick] (\xdelta,\aoneMax) -- (2*\xdelta,\atwoMax); 
			\ifnum \steps > 2
				\draw[\limitsLinestyle, color=POS_LIM_A, very thick] (2*\xdelta,\atwoMax) -- (3*\xdelta,\athreeMax); 
				\ifnum \steps > 3
					\draw[\limitsLinestyle, color=POS_LIM_A, very thick] (3*\xdelta,\athreeMax) -- (4*\xdelta,\afourMax);
					\fi
			\fi
		\fi
		\draw[\limitsLinestyle, color=POS_LIM_C, very thick] (0,\azeroMin) node[left=0.02cm, color=black] {$\boldsymbol{a_{t_{min}}}$} -- (\xdelta,\aoneMin); 
		\ifnum \steps > 1
			\draw[\limitsLinestyle, color=POS_LIM_C, very thick] (\xdelta,\aoneMin) -- (2*\xdelta,\atwoMin); 
			\ifnum \steps > 2
				\draw[\limitsLinestyle, color=POS_LIM_C, very thick] (2*\xdelta,\atwoMin) -- (3*\xdelta,\athreeMin); 
				\ifnum \steps > 3
					\draw[\limitsLinestyle, color=POS_LIM_C, very thick] (3*\xdelta,\athreeMin) -- (4*\xdelta,\afourMin);
				\fi
			\fi
		\fi
	\fi
	\ifnum \showRedLine = 1
	    \draw[\redLineLinestyle, color=POS_LIM_D, very thick] (0,\azeroMin)  -- (0,\azeroMax);
		\draw[\redLineLinestyle, color=POS_LIM_D, very thick] (\xdelta,\aoneMin)  -- (\xdelta,\aoneMax);
		\draw[\redLineLinestyle, color=POS_LIM_D, very thick] (2*\xdelta,\atwoMin)  -- (2*\xdelta,\atwoMax);
		\draw[\redLineLinestyle, color=POS_LIM_D, very thick] (3*\xdelta,\athreeMin)  -- (3*\xdelta,\athreeMax);
		\node[color=\actionColor, scale=0.9] at (0.5*\xdelta, \amin + 0.3) {$\underline{a}_t=0.0$}; 
		\node[color=\actionColor, scale=0.9] at (1.5*\xdelta, \amin + 0.3) {$\underline{a}_{t+1}=-0.5$}; 
		\node[color=\actionColor, scale=0.9] at (2.5*\xdelta, \amin + 0.3) {$\underline{a}_{t+2}=1.0$}; 
	\fi

	\ifnum \showAcc = 1
		\ifnum \violation = 0
			\draw[color=POS_LIM_B, very thick] (0,\azero)  -- (\xdelta,\aone) node[pos=1.0, above=0.13cm, color=black, xshift=0.6cm, name=a_t_plus_one_n, outer sep=0pt, inner sep=1.25pt]{$a_{t+1}$}; %
			\draw[-stealth] (a_t_plus_one_n.186) -- ($(\xdelta, \aone) + (0.03*\xdelta, 0.03*\xdelta)$);
			\fill[black] (\xdelta,\aone) circle (1.5pt);
    		\draw[|-stealth, thick, draw=\actionColor]  ($(\xdelta, \aoneMin) + (-0.065*\xdelta, 0)$) -- ($(\xdelta, \aone) + (-0.065*\xdelta, 0)$) node[pos=0.5, left=0.05cm, color=\actionColor, scale=0.9] {$\underline{a}_{t}$};
		\else
			\draw[color=POS_LIM_B, very thick, dashed] (0,\azero)  -- (\xdelta,\aoneViolation) node[pos=0, left=0.15cm, color=black] {$a_0$}; %
		\fi
		\ifnum \steps > 1
			\ifnum \violation = 0
				\draw[color=POS_LIM_B, very thick] (\xdelta,\aone) -- (2*\xdelta,\atwo) node[pos=1.0, above=-0.53cm, color=black, xshift=0.65cm, name=a_t_plus_two, outer sep=0pt, inner sep=1.75pt]{$a_{t+2}$};
				\draw[-stealth] (a_t_plus_two.174) -- ($(2*\xdelta, \atwo) + (0.03*\xdelta, -0.03*\xdelta)$);
    			\fill[black] (2*\xdelta,\atwo) circle (1.5pt);
        		\draw[|-stealth, thick, draw=\actionColor]  ($(2*\xdelta, \atwoMin) + (-0.065*\xdelta, 0)$) -- ($(2*\xdelta, \atwo) + (-0.065*\xdelta, 0)$) node[pos=0.65, left=0.05cm, color=\actionColor, scale=0.9] {$\underline{a}_{t+1}$};
			\else
				\draw[color=POS_LIM_B, very thick, dashed] (\xdelta,\aoneViolation) -- (2*\xdelta,\atwo); 
			\fi
			
			\ifnum \steps > 2
				\draw[color=POS_LIM_B, very thick] (2*\xdelta,\atwo) -- (3*\xdelta,\athree); 
				\fill[black] (3*\xdelta,\athree) circle (1.5pt);
        		\draw[|-stealth, thick, draw=\actionColor]  ($(3*\xdelta, \athreeMin) + (-0.065*\xdelta, 0)$) -- ($(3*\xdelta, \athree) + (-0.065*\xdelta, -0.2)$) node[pos=0.5, left=0.05cm, color=\actionColor, scale=0.9] {$\underline{a}_{t+2}$};
				\ifnum \steps > 3
					\draw[color=POS_LIM_B, very thick] (3*\xdelta,\athree) -- (4*\xdelta,\afour);
				\fi
			\fi
		\fi
	\fi

	\fill[black] (0,\azero) circle (1.5pt) node[above=\azero, left=0.12cm, color=black] {$a_t$};

	\draw[dotted, thick] (0, \amin) node[left=0.23cm] {$a_{min}$}  -- (3*\xdelta, \amin) ;
	\draw[dotted, thick] (0, \amax) node[left=0.16cm] {$a_{max}$}  -- (3*\xdelta, \amax) ;

	\end{tikzpicture} 
    } %
	\caption{%
	The figure illustrates for a single joint how actions $\underline{a}$ are mapped to kinematically feasible accelerations.
	}
	\vspace{-0.5cm}
	\label{fig:action_space}
\end{figure}
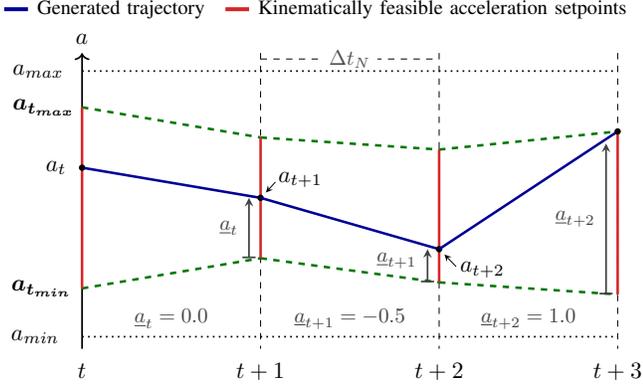
Each action $\underline{a} \in \mathcal{A}$ is a vector consisting of a scalar $\in [-1, 1]$ per robot joint controlled by the neural network. 

At time step $t$, the action $\underline{a_t}$ specifies the joint acceleration $a_{t+1}$.
Compliance with the specified joint limits is ensured by mapping the action $\underline{a_t}$ along the range of kinematically feasible acceleration setpoints $[a_{t+1_{min}}, a_{t+1_{max}}]$:
\begin{align}
a_{t+1} = a_{t+1_{min}} + \frac{1 + \underline{a}_t}{2} \cdot \left(a_{t+1_{max}} - a_{t+1_{min}}\right)
\end{align}
The method to compute the range $[a_{t+1_{min}}, a_{t+1_{max}}]$ is explained in \cite{kiemel2020learning}.
Fig. \ref{fig:action_space} demonstrates how the acceleration of a single joint is controlled using three exemplary actions $\underline{a_t}$, $\underline{a_{t+1}}$ and $\underline{a_{t+2}}$.  
To generate a continuous trajectory, the acceleration is linearly interpolated between the discrete time steps. 
Once the course of the joint acceleration is known, velocity and position setpoints for a trajectory controller can be calculated by integration. 

\subsubsection{Calculation of rewards $R_{\underline{a}}$}
We consider the tracking of reference paths as a multi-objective optimization problem.
The reward function formalizes the trade-off between the time to traverse the reference path, the deviation from the reference path and additional task-specific objectives:
\begin{align}
R_{\underline{a}} = \alpha \cdot R_l + \beta \cdot R_d + \gamma \cdot R_{s_1} + \ldots + \delta \cdot R_{s_n}, 
\label{eq:reward}
\end{align}
where $R_l$ is the path length reward, $R_d$ is the path deviation reward and $R_{s_1}$ to $ R_{s_n}$ represent task-specific objectives. %
The reward components $R_l$, $R_d$, $R_{s_1} \ldots R_{s_n}$ are defined to be in the range $[0.0, 1.0]$.
Non-negative weighting factors \mbox{$\alpha$, $\beta$, $\gamma \ldots \delta$} control the influence of each objective. 

To determine the path length reward $R_l$ and the path deviation reward $R_d$, we calculate the path traversed as a result of action $\underline{a}$. In the reward calculation box of Fig.~\ref{fig:basic_principle}, this path is shown in red.
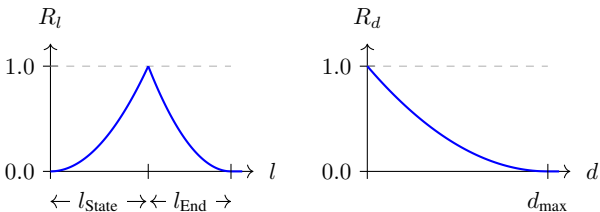
\begin{figure}[b]
\centering
\captionsetup[subfigure]{margin=0pt}
    \vspace{-0.2cm}
    \begin{subfigure}[c]{0.23\textwidth}
	   \vspace{-0.0cm}
	   \begin{tikzpicture}
  \def\legendScale{0.88}
  \def\tickLength{0.2}
  \def\lStateX{1.3}
  \def\lEndX{2.4}
  \def\yOne{1.4}
  \def\annotationOffset{0.4}
  \draw[->] (-0.1, 0) -- (2.7, 0) node[pos=0, left, scale=\legendScale]{$0.0$} node[right=0.1cm, scale=\legendScale] {$l$};
  \draw[->] (0, -0.1) -- (0, 1.7) node[above=0.1cm, scale=\legendScale] {$R_l$};
  \draw (-0.5*\tickLength, \yOne) -- + (\tickLength, 0) node[pos=0, left, scale=\legendScale]{$1.0$};
  \draw[dashed, draw=black!30] (0, \yOne) -- (\lEndX, \yOne);
  \draw[scale=1.0, domain=-0:\lStateX, smooth, variable=\x, blue, thick] plot ({\x}, {\x*\x*\yOne*(1/(\lStateX*\lStateX))});
  \draw[scale=1.0, domain=\lStateX:\lEndX, smooth, variable=\x, blue, thick] plot ({\x},
  {(\x-\lEndX)*(\x-\lEndX)*\yOne*(1/((\lStateX-\lEndX)*(\lStateX-\lEndX)))});
  \draw[blue, thick] (\lEndX, 0.0) -- +(0.15, 0);
  \draw (\lStateX, -0.5*\tickLength) -- + (0, \tickLength);
  \draw (\lEndX, -0.5*\tickLength) -- + (0, \tickLength);
  \draw[<->] (0, -1*\annotationOffset)  --  (\lStateX-0.025, -1*\annotationOffset) node[pos=0.5, fill=white, scale=\legendScale] {$l_{\textrm{State}}$};
  \draw[<->] (\lStateX+0.025, -1*\annotationOffset)  --  (\lEndX, -1*\annotationOffset) node[pos=0.5, fill=white, scale=\legendScale] {$l_{\textrm{End}}$};
  
\end{tikzpicture}

	\end{subfigure}
	\begin{subfigure}[c]{0.23\textwidth}
	    \vspace{-0.0cm}
	    \begin{tikzpicture}
  \def\legendScale{0.88}
  \def\tickLength{0.2}
  \def\lStateX{1.3}
  \def\lEndX{2.4}
  \def\yOne{1.4}
  \def\annotationOffset{0.4}
  \draw[->] (-0.1, 0) -- (2.7, 0) node[pos=0, left, scale=\legendScale]{$0.0$} node[right=0.1cm, scale=\legendScale] {$d$};
  \draw[->] (0, -0.1) -- (0, 1.7) node[above=0.1cm, scale=\legendScale] {$R_d$};
  \draw (-0.5*\tickLength, \yOne) -- + (\tickLength, 0) node[pos=0, left, scale=\legendScale]{$1.0$};
  \draw[dashed, draw=black!30] (0, \yOne) -- (\lEndX, \yOne);
  \draw[scale=1.0, domain=-0:\lEndX, smooth, variable=\x, blue, thick] plot ({\x}, {(\x-\lEndX)*(\x-\lEndX) *\yOne / (\lEndX*\lEndX)});
  \draw[blue, thick] (\lEndX, 0.0) -- +(0.15, 0);
  \draw (\lEndX, -0.5*\tickLength) -- + (0, \tickLength);
  \node[scale=\legendScale] at (\lEndX, -1*\annotationOffset){$d_{\textrm{max}}$}; 
  
\end{tikzpicture}

	\end{subfigure} 
	\vspace{-0.1cm}
	\caption{Path length reward $R_l$ and path deviation reward $R_d$.}
	\label{fig:rewards}
	\vspace{-0.1cm}
\end{figure}
We then compute the length of the path $l$ and its average deviation from the \mbox{reference path $d$}.
To calculate $d$, the Euclidian distance between waypoints on the generated path and waypoints on the reference path is averaged.
The first waypoint of the generated path is compared with the waypoint of the reference path that is marked with a red cross in Fig.~\ref{fig:observation_spline}. 
Further comparison points are shifted by the same arc length on each of the two paths.
Fig.~\ref{fig:rewards} illustrates how $l$ and $d$ are mapped to $R_l$ and $R_d$.
The path length reward $R_l$ controls the speed of traversing the reference path. 
Under normal conditions, a fast traversal is preferred. For that reason, larger path lengths receive higher rewards. 
However, towards the end of the reference path, the robot should slow down and finally come to a standstill.
To achieve this behavior, the reward is reduced if the path length exceeds $l_\text{State}$, the length of the path included in the state. %
Once the end of the reference spline is reached, $l_\text{State}$ is zero and the constant length $l_\text{End}$ controls how much the robot is penalized for further movements. The path deviation reward $R_d$ encourages the robot to stay close to the reference path. It is defined as a decreasing quadratic function yielding a value of zero if the deviation $d$ exceeds a predefined threshold $d_{\textrm{max}}$.

For the ball-on-plate task, an additional reward component $R_{s_1}$ is determined based on the distance between the ball and the center of the plate, using a decreasing quadratic function like the one shown for $R_d$.
To maintain balance with the bipedal humanoid \mbox{ARMAR-4}, we reward small angles between the robot's pelvis and an upward pointing z-axis. This angle is \SI{0}{\degree} when the robot is upright and \SI{90}{\degree} when the robot is lying on the ground. 
If the legs of the robot are not fixed, we additionally reward a small positional displacement of the pelvis to prevent the robot from moving around.

\vspace{0.1cm}
\subsubsection{Termination of episodes} In case of a successful task execution, an episode is terminated after a specified number of time steps. However, during training we define additional conditions that lead to an early termination of an episode. For path tracking without sensory feedback, an episode is terminated if the deviation $d$ between the generated path and the reference path exceeds a predefined threshold. 
The ball-on-plate task is additionally terminated if the ball falls from the plate. 
In case of the bipedal humanoid \mbox{ARMAR-4}, an episode is terminated if the robot falls over.
Note that the reward assigned to an action is never negative. 
For that reason, early termination leads to a lower sum of rewards and the neural network learns to avoid early termination during the training process.

\subsection{Generation of reference paths}

\subsubsection{Datasets used for the training process}
\begin{figure}[t]
\captionsetup[subfigure]{margin=0pt}
    \vspace{0.2cm}
    \hspace{-0.0025\textwidth}
    \begin{subfigure}[c]{0.155\textwidth}
	   \vspace{-0.0cm}
	   \begin{tikzpicture}[scale=1.0, every node/.style={scale=1.0}, node distance=2cm]
            \node[image_frame] at (0, 0.0cm) { \includegraphics[trim=925 300 900 350, clip, height=1.025\textwidth]{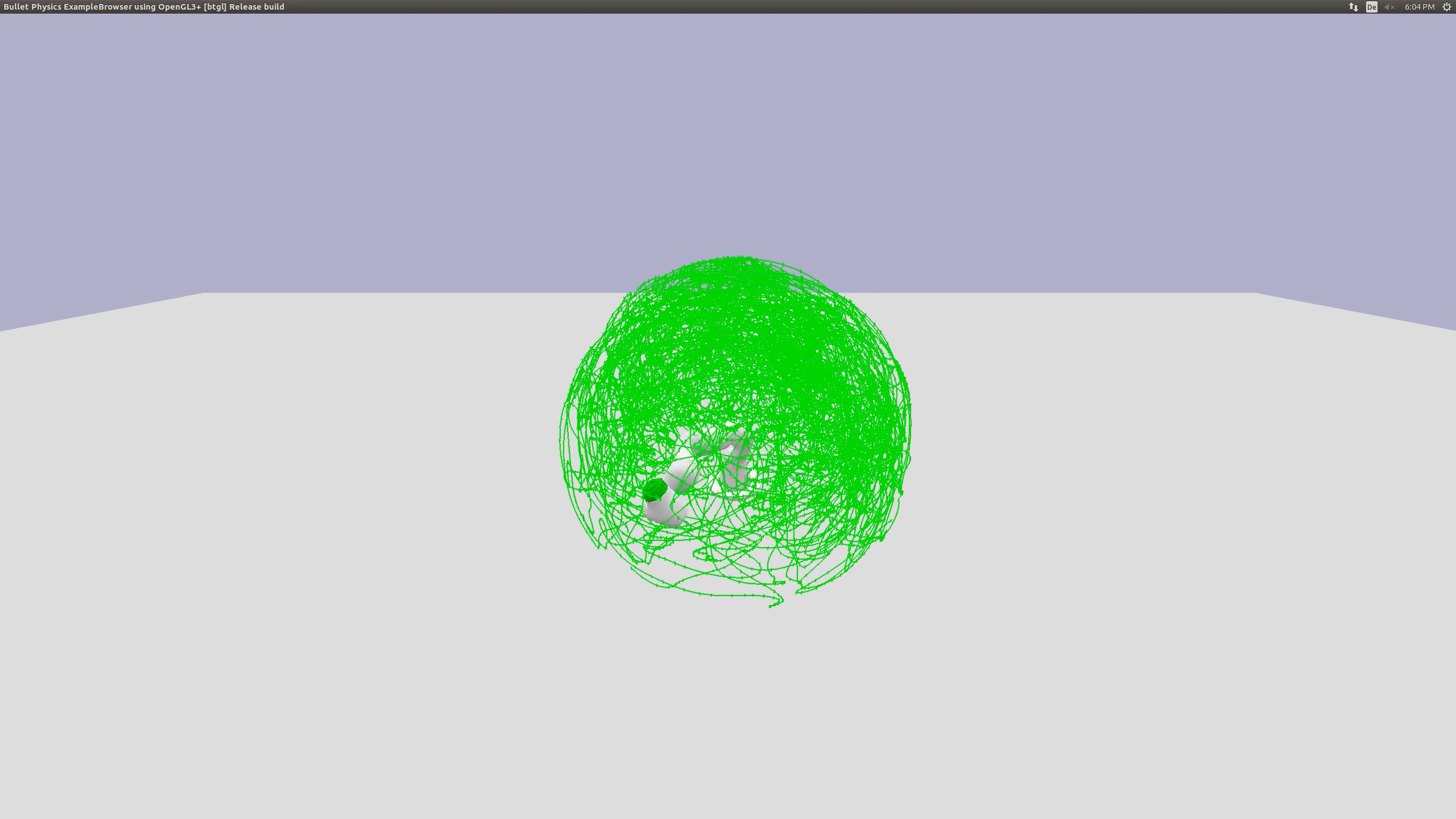}};
        \end{tikzpicture}
	  
	   \vspace{-0.40cm}\hspace*{-1.2cm}\subcaptionbox{Random}[5cm]

	\end{subfigure}
	\begin{subfigure}[c]{0.155\textwidth}
	    \vspace{-0.0cm}
	    \begin{tikzpicture}[scale=1.0, every node/.style={scale=1.0}, node distance=2cm]
            \node[image_frame] at (0, 0.0cm) { \includegraphics[trim=925 300 900 350, clip, height=1.025\textwidth]{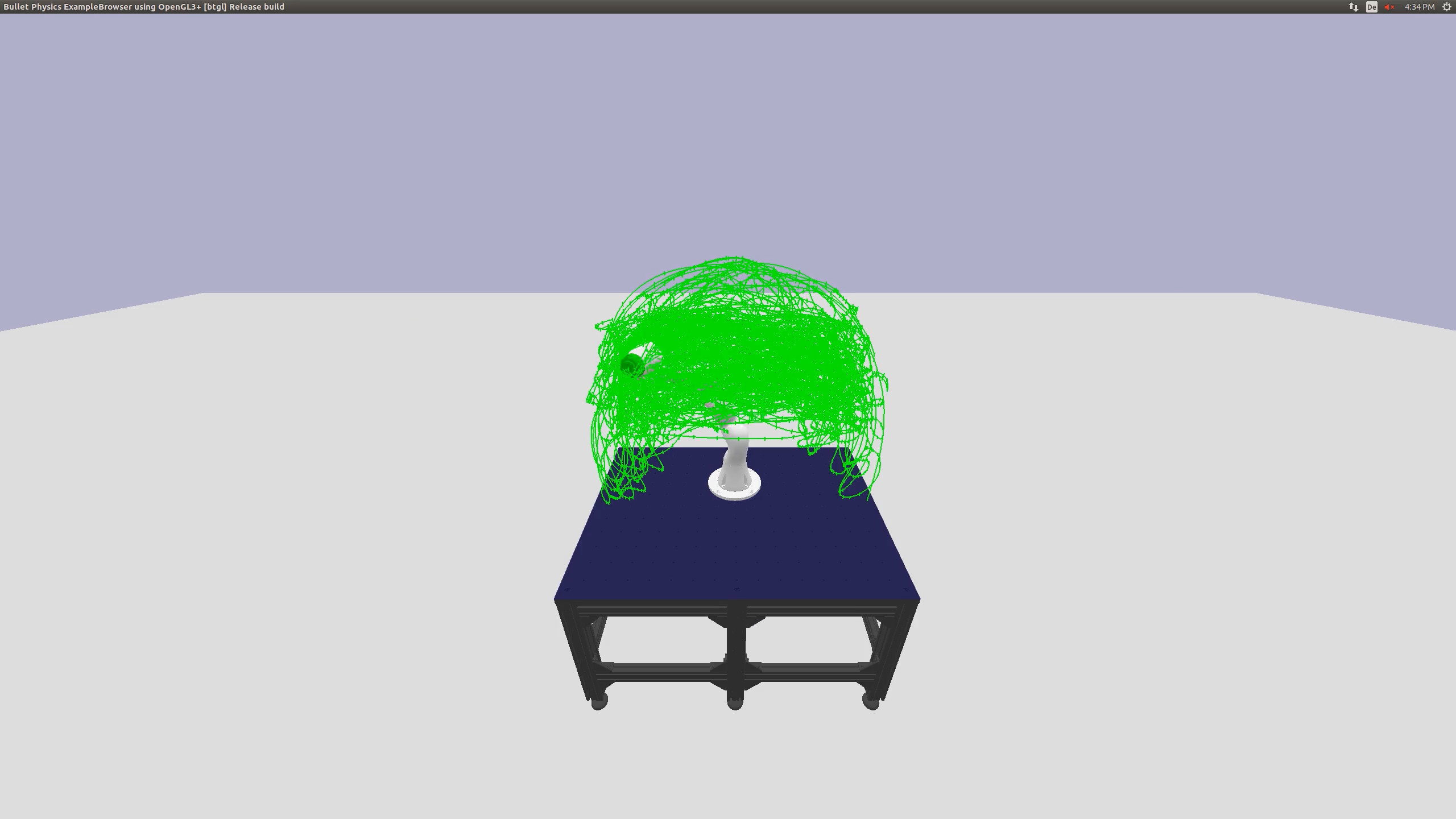}};
        \end{tikzpicture}

		\vspace{-0.4cm}\hspace*{-1.2cm}\subcaptionbox{Target point}[5cm]

	\end{subfigure} 
    \begin{subfigure}[c]{0.155\textwidth}
	   \vspace{0.0cm} 
	   \begin{tikzpicture}[scale=1.0, every node/.style={scale=1.0}, node distance=2cm]
            \node[image_frame] at (0, 0.0cm) { \includegraphics[trim=925 300 900 350, clip, height=1.025\textwidth]{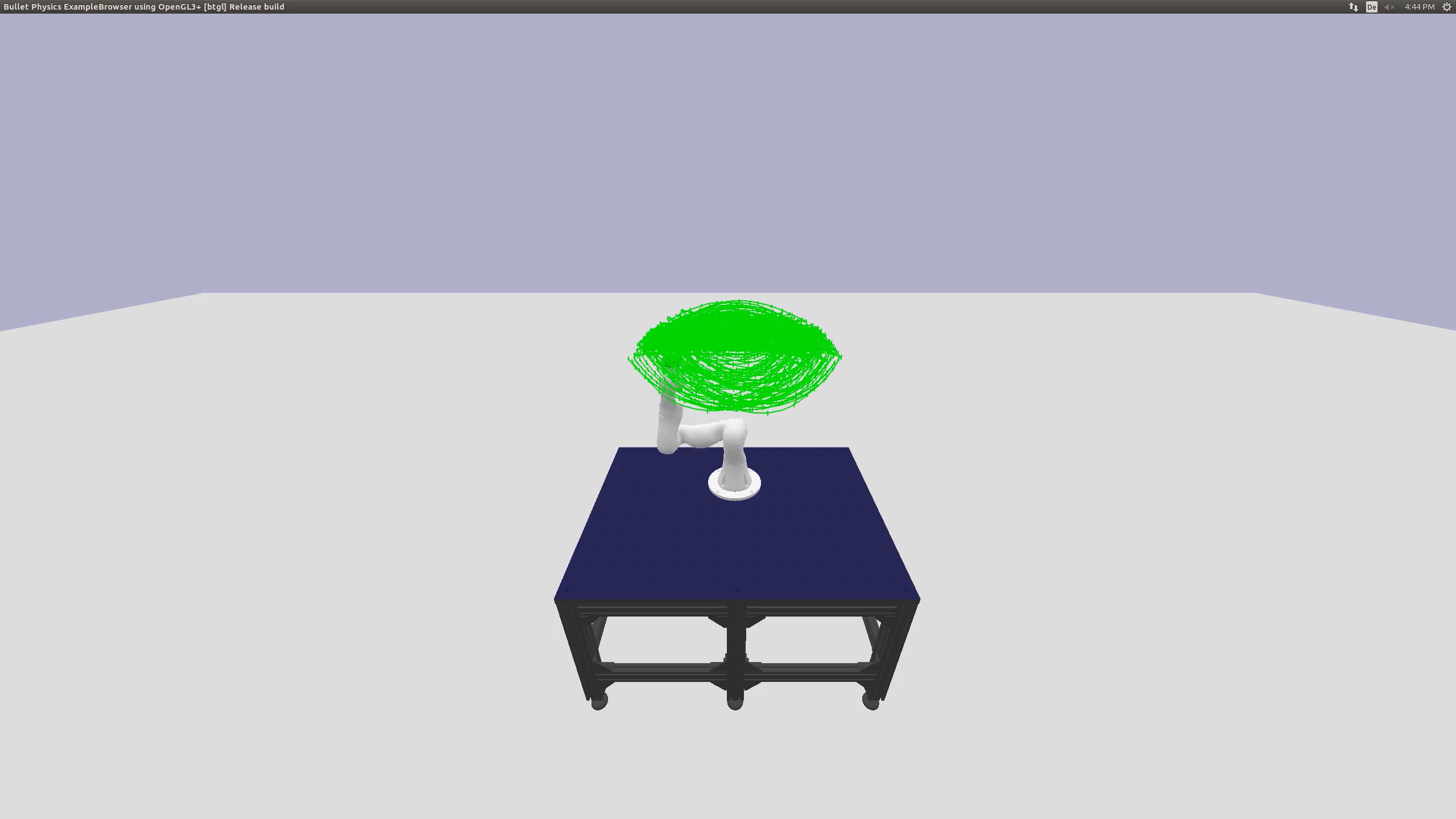}};
        \end{tikzpicture}

	   \vspace{-0.4cm}\hspace*{-1.2cm}\subcaptionbox{Ball balancing}[5cm]

	\end{subfigure}

	\caption{Visualization of the datasets used for our evaluation.}%
	\label{fig:datasets}
	\vspace{-0.28cm}
\end{figure}

When using model-free RL, the aim of the training process is to find a policy that optimizes the sum of future rewards.
Transferred to our specific problem, this means that the learning algorithm tries to find an optimized tracking strategy for the entire reference path, although only a part of it is known.
The best optimization results can be achieved if the reference paths used within the training process are similar to those encountered during deployment. 
For our evaluation, we use three datasets with different path characteristics. 
Fig.~\ref{fig:datasets} visualizes some of the reference paths for each dataset.

The random dataset is generated by selecting random actions at each decision step and storing the paths resulting from the random robot motions. We use a method described in \cite{kiemel2021collision} to ensure that only collision-free paths are generated. As can be seen in Fig.~\ref{fig:datasets}a, the reference paths cover the entire working space of the robot. 
In typical industrial applications, however, robots do not move randomly. Instead, they often move between Cartesian target points. 
To recreate such a situation, we train a neural network to generate collision-free trajectories between randomly sampled target points using the method from \cite{kiemel2021collision}. 
The target point dataset shown in Fig.~\ref{fig:datasets}b is composed of paths produced by this network. 
For the ball-on-plate task, the reference paths are computed in such a way that the plate is always aligned horizontally. The corresponding dataset is visualized in Fig.~\ref{fig:datasets}c.
We note that the datasets used for training and testing are created in the same way, but do not contain the same paths. 

\subsubsection{Spline knots included in the state}

\begin{figure}[h]
\captionsetup[subfigure]{margin=0pt}
    \vspace{-0.5cm}
    \centering
    \begin{tikzpicture}
	        \definecolor{ORIGINAL_PATH_BLUE}{RGB}{0, 0, 255}
	        \definecolor{MATPLOTLIB_ORANGE}{RGB}{255, 127, 14}
	        \definecolor{MATPLOTLIB_DIMGREY}{RGB}{105, 105, 105}
            \hspace{0.23cm}
             \node[text width=4cm] at (0.2cm, 4.35cm) (origin){};
            	\draw [draw=ORIGINAL_PATH_BLUE, line width=1.5pt, scale=0.7] ($(origin.center)+(-9.3cm, -0.0cm)$) -- + (0.45cm, 0cm) node[pos=1, right, yshift=0.01cm, align=left, scale=0.85]{Original path\strut};
            	\draw [draw=black, line width=0.9pt, scale=0.7] ($(origin.center)+(-6.0cm, -0.075cm)$) -- + (0.2cm, 0.2cm) node[pos=1, right, yshift=0.01cm, align=left, scale=0.85]{};
            	\draw [draw=black, line width=0.9pt, scale=0.7] ($(origin.center)+(-6.0cm, +0.125cm)$) -- + (0.2cm, -0.2cm) node[pos=1, right, yshift=0.01cm, align=left, scale=0.85]{};
            	\node[align=left, scale=0.85] at  ($(origin.center)+(-3.12cm, -0.0cm)$) {Spline knots\strut};
            	\draw [draw=REFERENCE_PATH_GREEN, line width=1.5pt, scale=0.7] ($(origin.center)+(-3.0cm, -0.0cm)$) -- + (0.45cm, 0cm) node[pos=1, right, yshift=0.01cm, align=left, scale=0.85]{Resulting reference path\strut};
    \end{tikzpicture}
    \begin{subfigure}[c]{0.23\textwidth}
	   \vspace{-0.0cm}
	   \begin{tikzpicture}[scale=1.0, every node/.style={scale=1.0}, node distance=2cm]
            \node[image_frame](eight_equal_sampling) at (0, 0.0cm) {\includegraphics[trim=200 970 270 950, clip, height=0.6\textwidth]{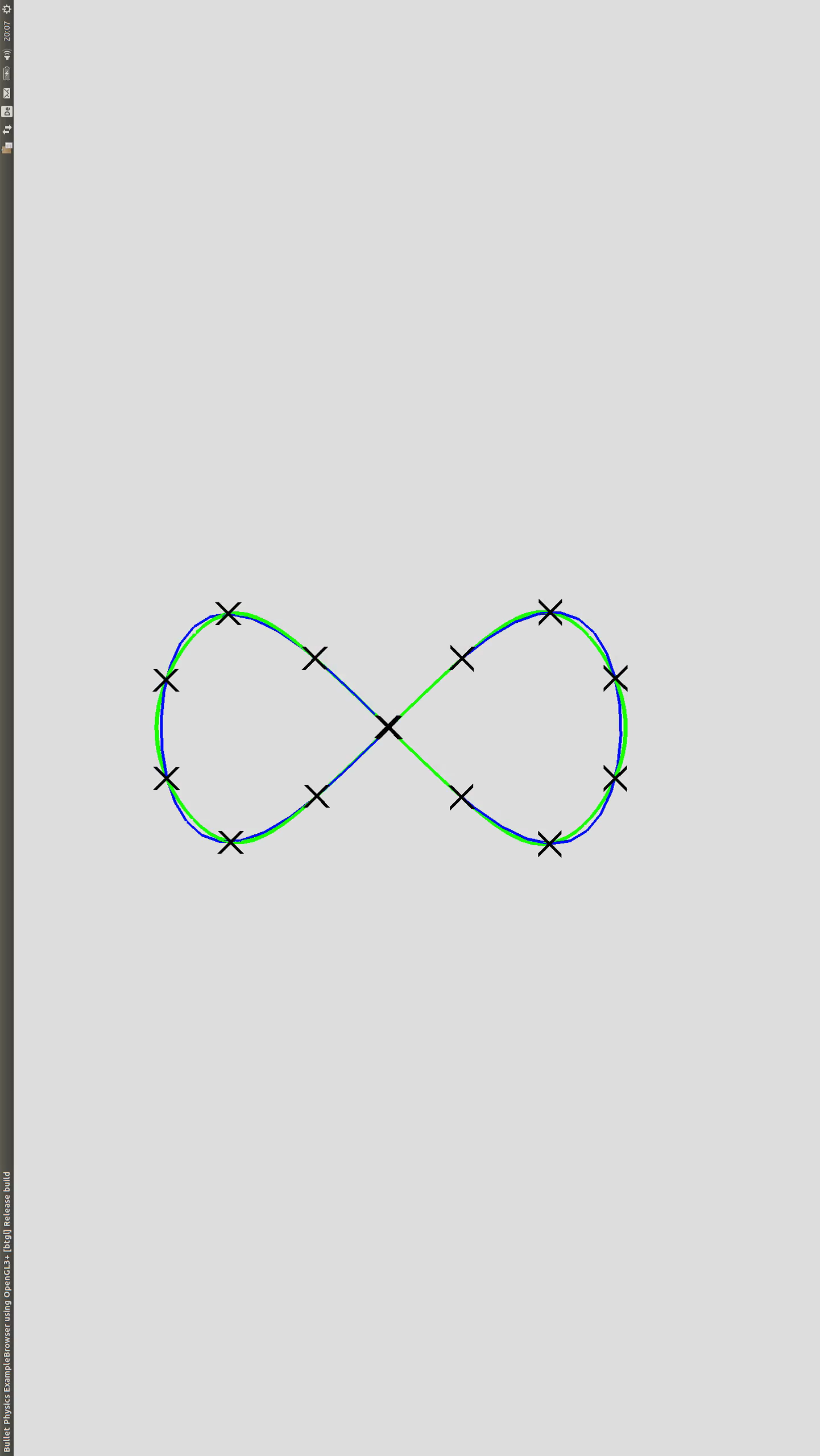}};
       \end{tikzpicture}

	   \vspace{-0.40cm}\hspace*{-0.65cm}\subcaptionbox{Distance-based sampling}[5cm]

	\end{subfigure}
	\begin{subfigure}[c]{0.23\textwidth}
	    \vspace{-0.0cm}
	    \begin{tikzpicture}[scale=1.0, every node/.style={scale=1.0}, node distance=2cm]
            \node[image_frame](eight_curvature) at (0, 0.0cm) {\includegraphics[trim=200 970 270 950, clip, height=0.6\textwidth]{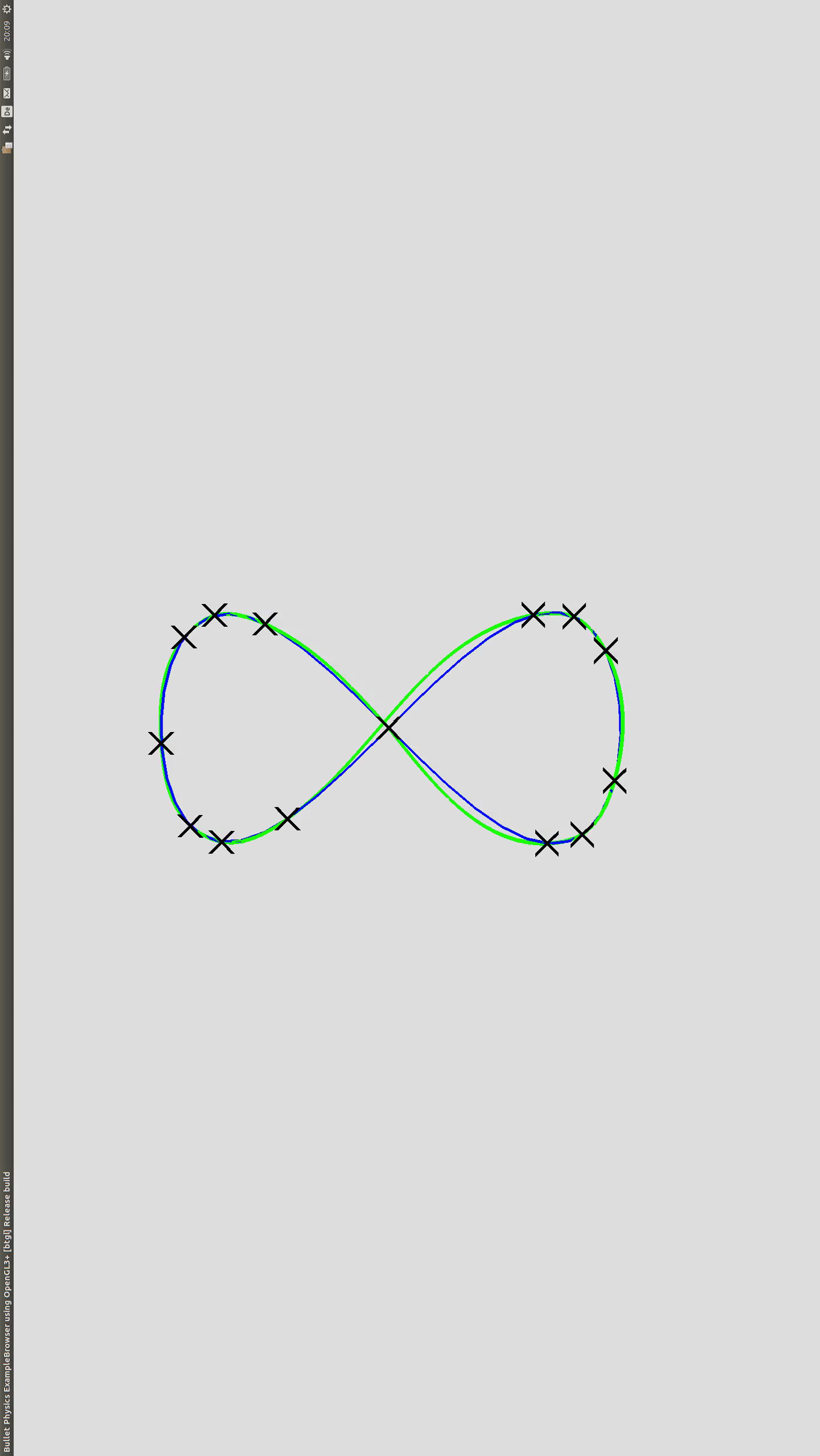}};
       \end{tikzpicture}

		\vspace{-0.4cm}\hspace*{-0.645cm}\subcaptionbox{Curvature-based sampling}[5cm]

	\end{subfigure} 
	
	\begin{subfigure}[c]{0.475\textwidth}
	    \vspace{0.1cm}
	    \begin{tikzpicture}
	        \definecolor{MATPLOTLIB_BLUE}{RGB}{31, 119, 180}
	        \definecolor{MATPLOTLIB_ORANGE}{RGB}{255, 127, 14}
	        \definecolor{MATPLOTLIB_DIMGREY}{RGB}{105, 105, 105}
            \hspace{0.23cm}
             \node[text width=4cm] at (0.2cm, 4.35cm) (origin){};
            	\draw [draw=MATPLOTLIB_BLUE, line width=1.5pt, scale=0.7] ($(origin.center)+(-9.3cm, -0.0cm)$) -- + (0.45cm, 0cm) node[pos=1, right, yshift=0.01cm, align=left, scale=0.85]{$x(s)$};
            	\draw [draw=MATPLOTLIB_ORANGE, line width=1.5pt, scale=0.7] ($(origin.center)+(-7.4cm, -0.0cm)$) -- + (0.45cm, 0cm) node[pos=1, right, yshift=0.01cm, align=left, scale=0.85]{$y(s)$};
        	    \draw [draw=MATPLOTLIB_DIMGREY, line width=1.5pt, scale=0.7] ($(origin.center)+(-5.5cm, 0.0cm)$) -- + (0.45cm, 0cm) node[pos=1, right, yshift=0.00cm, align=left, scale=0.85]{Curvature $\kappa(s)=\sqrt{x''(s)^2 + y''(s)^2}$};
         \end{tikzpicture}
        \vspace{0.1cm}
	    \includegraphics[trim=8 62 0 0, clip, width=\textwidth]{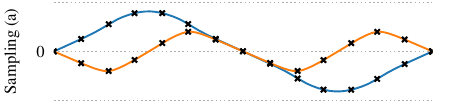}
	    \includegraphics[trim=8 60 0 0, clip, width=\textwidth]{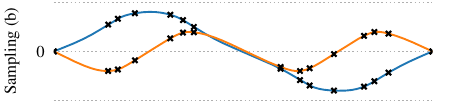}
	    \includegraphics[trim=8 0 0 54, clip, width=\textwidth]{figures/spline_sampling/eight_equal_pos_curvature.pdf}

		\vspace{-0.4cm}\hspace*{-0.845cm}\subcaptionbox{Knot position over arc length for both sampling strategies.}[10cm]
		
	\end{subfigure}

	\caption{%
	Strategies for the sampling of spline knots.}%
	\label{fig:spline_sampling}
	\vspace{-0.28cm}
\end{figure}

The desired reference path is defined based on a fixed number of spline knots included in the state.  
Different strategies can be used to place the knots along a given path. 
Fig. \ref{fig:spline_sampling} illustrates two of them using a two-dimensional path shaped like a lemniscate. 
With distance-based sampling, the knots are placed so that the arc length between them is equal. 
When using curvature-based sampling, the curvature of the path is computed and integrated. 
The knots are selected such that the integrated curvature between the knots is equal. 
With distance-based sampling, the length of the path included in the state is always the same, except at the end of the reference path. When using curvature-based sampling, however, the included path length depends on the curvature of the reference path. 
Sections of low curvature that can be traversed more quickly lead to a longer path being described by the state.

\section{Evaluation}

We evaluate our method by learning to track reference paths with and without additional objectives. 
Our evaluation environments are shown in Fig. \ref{fig:header}.
The KUKA iiwa is an industrial lightweight robot with 7 joints. 
The humanoid robot \mbox{ARMAR-6}, shown in Figure \ref{fig:header}c, is controlled by 17~joints. 
The bipedal humanoid ARMAR-4 has 30 joints, with 18 used for the upper body and 12 used for the legs.
In the case of ARMAR-4, the reference path specifies the joint positions of the upper body only. 
For the other two robots, the positions of all joints are defined by the reference path.

Our neural networks are trained using proximal policy optimization (PPO) \cite{schulman2017proximal} based on data generated by the physics engine PyBullet\cite{coumans2016pybullet}.
We use networks with two hidden layers, the first one consisting of 256 neurons and the second one consisting of 128 neurons. 
The time between decision steps is set to $\Delta t_{N}=$ \SI{0.1}{\s}.
The results shown in the following tables were obtained by averaging data from 1200 episodes, with reference paths taken from separate test datasets. 
For the random dataset and the target point dataset, we use curvature-based sampling with $N=9$ knots included in the state.
In the case of the ball balancing dataset, distance-based sampling with $N=5$ knots is used. 

\begin{figure}[b]
\captionsetup[subfigure]{margin=0pt}
    \vspace{-0.3cm}
    \centering
        \begin{tikzpicture}
                \def\xminPlot{0.151} 
                \def\xmaxPlot{0.949} 
                \SUBTRACT{\xmaxPlot}{\xminPlot}{\xdeltaPlot}
                \DIVIDE{\xdeltaPlot}{3}{\xdeltaPlotNorm}
                \def\yminPlot{0.135}
                \def\ymaxPlot{0.985}
                \def\yminAcc{0.364}
                \def\ymaxAcc{0.535}
                \SUBTRACT{\ymaxPlot}{\yminPlot}{\ydeltaPlot}
                \SUBTRACT{\ymaxAcc}{\yminAcc}{\ydeltaAcc}
                \MULTIPLY{\ydeltaAcc}{4}{\ydeltaGraph}
                \SUBTRACT{\ydeltaPlot}{\ydeltaGraph}{\ydeltaTmp}
                \DIVIDE{\ydeltaTmp}{3}{\ydeltaGap}
                
                 \node[text width=4cm] at (-1.2cm, 0cm) (origin){};
                	\draw [draw=REFERENCE_PATH_GREEN, line width=1.5pt, scale=0.7] ($(origin.center)+(-6.75cm, 0.075cm)$) -- + (0.45cm, 0cm) node[pos=1, right, yshift=-0.05cm, align=left, scale=0.85]{Reference path};
                	\draw [draw=REFERENCE_PATH_PURPLE, line width=1.5pt, scale=0.7] ($(origin.center)+(-6.75cm, -0.0cm)$) -- + (0.45cm, 0cm);
                	
            	    \draw [draw=GENERATED_PATH_RED, line width=1.5pt, scale=0.7] ($(origin.center)+(-0.75cm, 0.0cm)$) -- + (0.45cm, 0cm) node[pos=1, right, yshift=0.00cm, align=left, scale=0.85]{Generated path};
    \end{tikzpicture}
    
    \begin{subfigure}[c]{0.23\textwidth}
	   \vspace{-0.0cm}
	   \begin{tikzpicture}[scale=1.0, every node/.style={scale=1.0}, node distance=2cm]
            \node[image_frame] at (0, 0.0cm) { \includegraphics[trim=855 420 860 250, clip, height=0.85\textwidth]{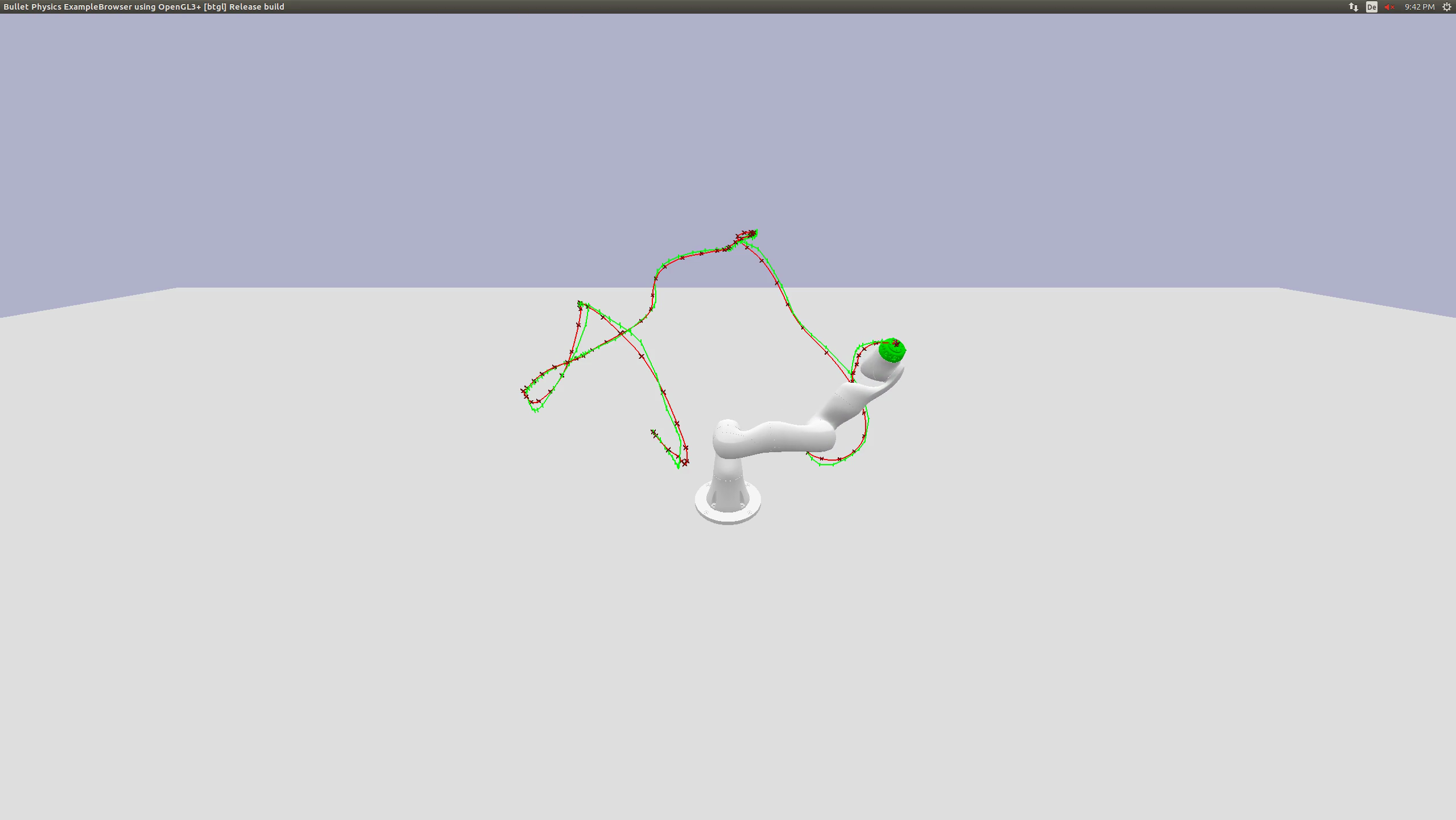}};
       \end{tikzpicture}

	   \vspace{-0.40cm}\hspace*{-0.7cm}\subcaptionbox{KUKA iiwa}[5cm]

	\end{subfigure}
	\begin{subfigure}[c]{0.23\textwidth}
	    \vspace{-0.0cm}
	    \begin{tikzpicture}[scale=1.0, every node/.style={scale=1.0}, node distance=2cm]
            \node[image_frame] at (0, 0.0cm) { \includegraphics[trim=800 390 800 170, clip, height=0.85\textwidth]{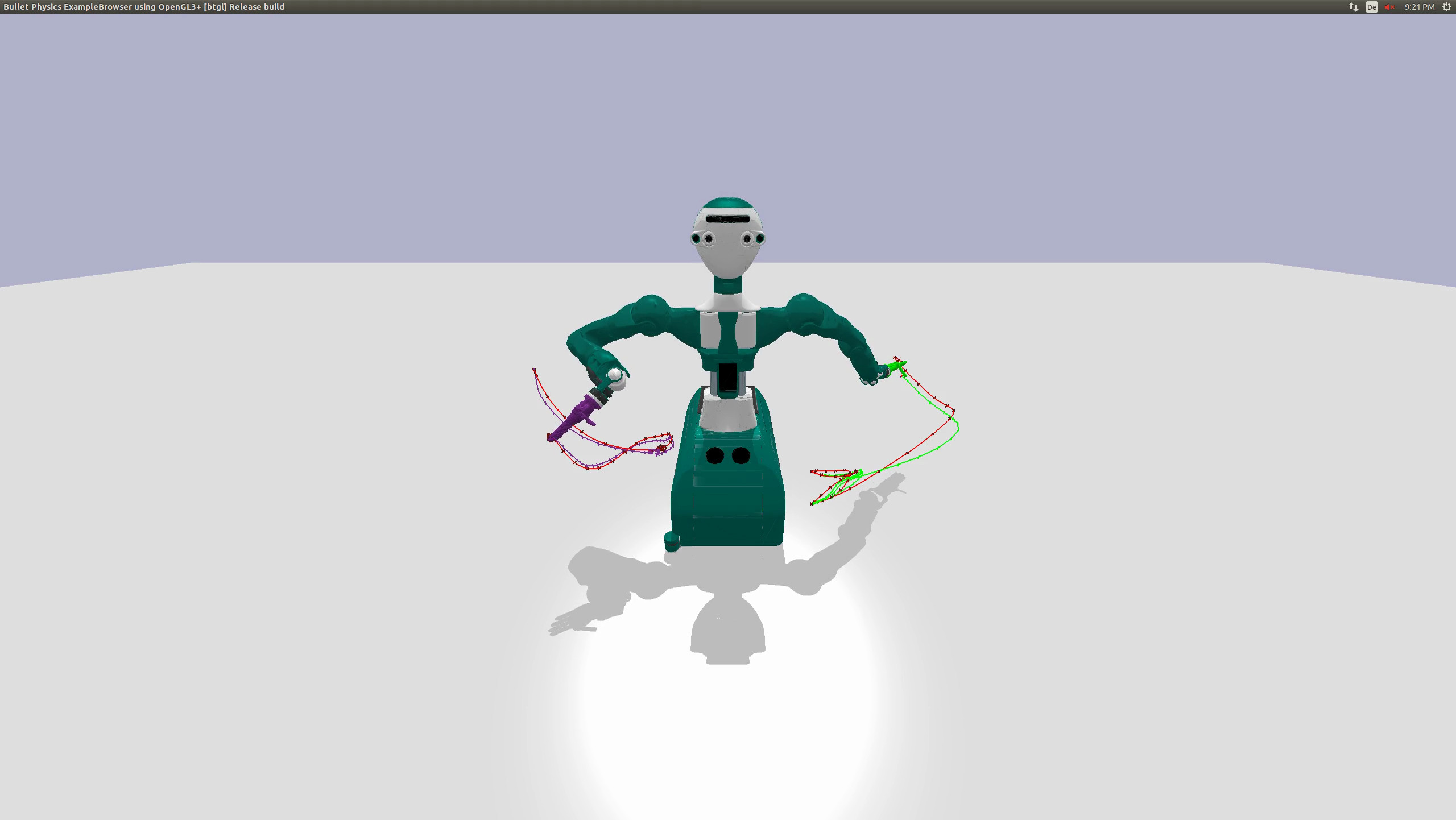}};
        \end{tikzpicture}

		\vspace{-0.4cm}\hspace*{-0.645cm}\subcaptionbox{ARMAR-6}[5cm]

	\end{subfigure}

	\caption{%
	Path tracking without additional objectives.}%
	\label{fig:example_path_no_feedback}
	\vspace{-0.28cm}
\end{figure}
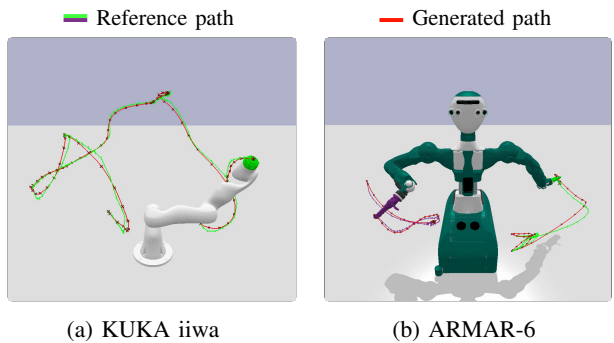

\begin{table*}[t]
    \caption{Training results for time-optimized path tracking without additional objectives obtained based on 1200 episodes.}
    \vspace{-0.15cm}
    \makegapedcells
\begin{tabular*}{\textwidth}{p{34.5mm}p{19mm}p{8.7mm}p{8.7mm}p{8.7mm}p{8.7mm}p{8.7mm}p{8.7mm}p{8.7mm}p{8.7mm}p{8.7mm}} 
    \toprule

\multirow[t]{2}{*}{Configuration}
     & \multicolumn{1}{c}{Duration [s]} & \multicolumn{3}{c}{Joint position deviation [rad]} & \multicolumn{3}{c}{Cart. position deviation [cm]}  & \multicolumn{3}{c}{Cart. orientation deviation [°]}\\
     & (robot stopped) & \multicolumn{1}{c}{mean} & \multicolumn{1}{c}{max}  & \multicolumn{1}{c}{final} & \multicolumn{1}{c}{mean} & \multicolumn{1}{c}{max}  & \multicolumn{1}{c}{final} & \multicolumn{1}{c}{mean} & \multicolumn{1}{c}{max}  & \multicolumn{1}{c}{final}\\
    \hline
KUKA iiwa & & & & \\
\tabitem Random dataset
    & \hfil $4.28$ \hfil & \hfil $ 0.11$ \hfil   &  \hfil $0.19$ \hfil   &  \hfil $0.09$ \hfil    &  \hfil $3.3$ \hfil  & \hfil $7.5$ \hfil  & \hfil $2.7$ \hfil  & \hfil $5.8$ \hfil  &  \hfil $11.8$ \hfil  & \hfil $5.1$ \hfil  \\
\tabitem Target point dataset %
   & \hfil $4.99$ \hfil & \hfil $0.12$ \hfil   &  \hfil $0.21$ \hfil   &  \hfil $0.12$ \hfil    &  \hfil $3.7$ \hfil  & \hfil $8.1$ \hfil  & \hfil $3.8$ \hfil  & \hfil $6.7$ \hfil  &  \hfil $13.5$ \hfil  & \hfil $6.3$ \hfil  \\ 
\tabitem Ball balancing dataset
    & \hfil $2.44$ \hfil & \hfil $0.04$ \hfil   &  \hfil $0.08$ \hfil   &  \hfil $0.03$ \hfil    &  \hfil $1.4$ \hfil  & \hfil $3.0$ \hfil  & \hfil $1.4$ \hfil  & \hfil $1.7$ \hfil  &  \hfil $3.9$ \hfil  & \hfil $1.5$ \hfil  \\
     \hline
ARMAR-6 & & & & \\
\tabitem Random dataset
    & \hfil $4.98$ \hfil & \hfil $0.14$ \hfil   &  \hfil $0.20$ \hfil   &  \hfil $0.16$ \hfil    &  \hfil $5.5$ \hfil  & \hfil $13.6$ \hfil  & \hfil $6.2$ \hfil  & \hfil $5.3$ \hfil  &  \hfil $11.5$ \hfil  & \hfil $6.0$ \hfil  \\
    \hline %
ARMAR-4 (fixed base)  & & & & \\  
\tabitem Random dataset
   & \hfil $5.09$ \hfil & \hfil $0.14$ \hfil   &  \hfil $0.20$ \hfil   &  \hfil $0.14$ \hfil    &  \hfil $3.3$ \hfil  & \hfil $7.8$ \hfil  & \hfil $3.5$ \hfil  & \hfil $5.6$ \hfil  &  \hfil $11.7$ \hfil  & \hfil $5.7$ \hfil  \\
\tabitem Target point dataset
    & \hfil $5.48$ \hfil & \hfil $0.14$ \hfil   &  \hfil $0.21$ \hfil   &  \hfil $0.15$ \hfil    &  \hfil $3.6$ \hfil  & \hfil $8.8$ \hfil  & \hfil $3.8$ \hfil  & \hfil $5.6$ \hfil  &  \hfil $12.4$ \hfil  & \hfil $6.2$ \hfil  \\
    \bottomrule
    \end{tabular*}
\label{table:tracking_no_feedback}
\vspace{-0.33cm}
\end{table*}

\subsection{Path tracking without additional objectives}
When tracking paths without additional targets, the networks are trained to minimize the duration of the trajectory and the average  deviation from the reference path. 

TABLE \ref{table:tracking_no_feedback} shows the training results obtained for the different robots and datasets. 
Note that the learning algorithm tries to minimize the average joint position deviation, given in the second column.  
However, to give a better idea of the results, we also specify deviations with respect to the position and orientation in Cartesian space. 
For this purpose, the reference path and the generated path are converted to Cartesian space using forward kinematics. 
The tool center point (TCP) is used as the reference point for the KUKA~iiwa, whereas the fingertips are used for the humanoid robots. 
Renderings of two exemplary episodes can be seen in Fig. \ref{fig:example_path_no_feedback}.
To compute the deviations, points on the reference path are compared with points on the generated path.
The comparison points are selected such that they are equally far away from the beginning of the respective path.
To specify a trajectory duration and a final position deviation, the robot joints are decelerated as soon as the end of the reference path is reached. 
Using the ball balancing dataset, an average Cartesian deviation of \SI{1.4}{\cm} and \SI{1.7}{\degree} is obtained.
The paths in the random  and the target point dataset show a larger variance, resulting in a Cartesian deviation of approximately \SI{4}{\cm} and~\SI{6}{\degree}.
In the following, we evaluate how the trajectory duration and the position deviation can be influenced.
\subsubsection{Trade-off traversing time vs. path deviaton}

\begin{table}[t]
    \vspace{-0.0cm}
    \caption{%
    Trade-off traversing time vs. path deviation.
    }
    \vspace{-0.15cm}
    \makegapedcells
\begin{tabular*}{0.49\textwidth}{@{}p{27.0mm}p{17.5mm}p{8.7mm}p{8.7mm}p{8.7mm}} 
    \toprule
\hspace{0.05cm}KUKA iiwa with & \multicolumn{1}{c}{Duration [s]} & \multicolumn{3}{c}{Cart. position deviation [cm]} \\
\hspace{0.05cm}random dataset  & (robot stopped) & \multicolumn{1}{c}{mean} & \multicolumn{1}{c}{max}  & \multicolumn{1}{c}{final} \\
 \hline
\hspace{0.05cm}\tabitem Length $<$ deviation
    &  \hfil $4.76$ \hfil  &   \hfil $2.7$ \hfil   &  \ \hfil $6.4$ \hfil &  \hfil $2.3$ \hfil\\
\hspace{0.05cm}\tabitem Length $\approx$ deviation
    &  \hfil $4.28$ \hfil  &   \hfil $3.3$ \hfil   &  \ \hfil $7.5$ \hfil &  \hfil $2.7$ \hfil\\
\hspace{0.05cm}\tabitem Length $>$ deviation
    &  \hfil $4.08$ \hfil  &   \hfil $3.6$ \hfil   &  \ \hfil $8.0$ \hfil &  \hfil $3.1$ \hfil\\

    \bottomrule
    \end{tabular*}
    \vspace{-0.54cm}
    
\label{table:trade_off}
\end{table}
Our reward function (\ref{eq:reward}) allows us to assign different weights to the path length reward $R_l$ and to the path deviation reward $R_d$.
TABLE \ref{table:trade_off} shows how the weighting affects the traversing time and the path deviation. 
If the weighting of the deviation reward is increased, the path deviation decreases whereas the trajectory duration increases. 
Likewise, faster trajectories with a higher path deviation are learned when the weighting of the path length reward is increased.

\subsubsection{Number of knots included in the state}
TABLE \ref{table:knots} shows how the training results are affected by the number of knots $N$ used to describe the following part of the reference path. If more knots are included in the state, the reference path can be traversed faster. In return, however, a larger part of the reference path has to be known in advance.  
\begin{table}[t]
    \vspace{-0.0cm}
    \caption{%
    Impact of the knots included in the state. 
    }
    \vspace{-0.15cm}
    \makegapedcells
\begin{tabular*}{0.49\textwidth}{p{22.5mm}p{19mm}p{8.7mm}p{8.7mm}p{8.7mm}} 
    \toprule
KUKA iiwa with & \multicolumn{1}{c}{Duration [s]} & \multicolumn{3}{c}{Cart. position deviation [cm]} \\
random dataset  & (robot stopped) & \multicolumn{1}{c}{mean} & \multicolumn{1}{c}{max}  & \multicolumn{1}{c}{final} \\
    \hline

\tabitem 5 knots
    &  \hfil $5.20$ \hfil  &   \hfil $2.5$ \hfil   &  \ \hfil $5.6$ \hfil &  \hfil $2.8$ \hfil\\

\tabitem 7 knots
    &  \hfil $4.52$ \hfil  &   \hfil $3.4$ \hfil   &  \ \hfil $7.5$ \hfil &  \hfil $3.0$ \hfil\\

\tabitem 9 knots
    &  \hfil $4.28$ \hfil  &   \hfil $3.3$ \hfil   &  \ \hfil $7.5$ \hfil &  \hfil $2.7$ \hfil\\

    \bottomrule
    \end{tabular*}
    \vspace{-0.5cm}
    
\label{table:knots}
\end{table}
\begin{table}[b]
    \vspace{-0.25cm}
    \caption{%
    Generalization ability between datasets.
    }
    \vspace{-0.15cm}
    \makegapedcells
\begin{tabular*}{0.49\textwidth}{@{}p{27.0mm}p{17.5mm}p{8.5mm}p{7.5mm}p{7.5mm}} 
    \toprule
\hspace{0.05cm} & \multicolumn{1}{c}{Duration [s]} & \multicolumn{3}{c}{Cart. position deviation [cm]} \\
\hspace{0.05cm}  & (robot stopped) & \hfil mean \hfil & \hfil max \hfil  & \hfil final \hfil \\
 \hline
\hspace{0.05cm}KUKA iiwa & & & & \\
\hspace{0.05cm}\tabitem Random dataset to \newline \hspace*{0.05cm}\phantom{\tabitem }target point dataset
    &  \hfil $4.87$ \hfil  &   \hfil $3.9$ \hfil   &  \hfil $9.2$ \hfil &  \hfil $2.7$ \hfil\\
\hspace{0.05cm}ARMAR-4 (fixed base) & & & & \\
\hspace{0.05cm}\tabitem Random dataset to \newline \hspace*{0.05cm}\phantom{\tabitem }target point dataset
    &  \hfil $5.40$ \hfil  &   \hfil $3.8$ \hfil   &   \hfil $9.2$ \hfil &  \hfil $4.1$ \hfil\\
    \bottomrule
    \end{tabular*}
    \vspace{-0.1cm}
    
\label{table:random_to_target}
\end{table}
\begin{table}[b]
     \vspace{-0.1cm}
     \caption{Feature comparison with TOPP-RA \cite{pham2018Toppra}}
     \vspace{-0.15cm}
    \makegapedcells
\begin{tabular*}{0.49\textwidth}{p{44mm}p{15.5mm}p{15.5mm}} 
    \toprule
\hspace{0.02cm}  & \multicolumn{1}{c}{TOPP-RA}  & \multicolumn{1}{c}{Ours} \\
    \hline
Supports velocity limits & \hfil \yes \hfil  & \hfil \yes \hfil \\
Supports acceleration limits & \hfil \yes \hfil  & \hfil \yes \hfil \\
Supports jerk limits & \hfil \no \hfil  & \hfil \yes \hfil \\
Supports path adjustments & \hfil \no \hfil  & \hfil \yes \hfil \\
Supports sensory feedback & \hfil \no \hfil  & \hfil \yes \hfil \\
    \bottomrule
    \end{tabular*}
    \vspace{0ex}
\label{table:feature_toppra}
\vspace{-0.0cm}
\end{table}
\vspace{0.1cm}
\subsubsection{Generalization ability between datasets}
To analyze the generalization ability with respect to different path characteristics,
we evaluate networks trained using the random dataset with reference paths from the target point dataset.
TABLE \ref{table:random_to_target} shows the results for the KUKA iiwa and the humanoid \mbox{ARMAR-4}.
Compared to the networks trained directly with the target point dataset, the resulting trajectories are slightly faster but 
also a little less accurate in tracking the reference path. 
Overall, however, the differences are small, indicating that networks trained on random paths can also be used to track paths with different path characteristics. 

In the accompanying video\href{\linkToVideo}{$^3$}, we additionally show how a network trained on random paths performs on reference paths that resemble geometric shapes in Cartesian space. 

\vspace{0.05cm}
\subsubsection{Comparison with time-optimal path parameterization}
We benchmark our approach with TOPP-RA \cite{pham2018Toppra}, a state-of-the-art offline method for time-optimal path parameterization.
TABLE \ref{table:feature_toppra} compares the features of both methods. 
 
\begin{figure*}[t]
    \input{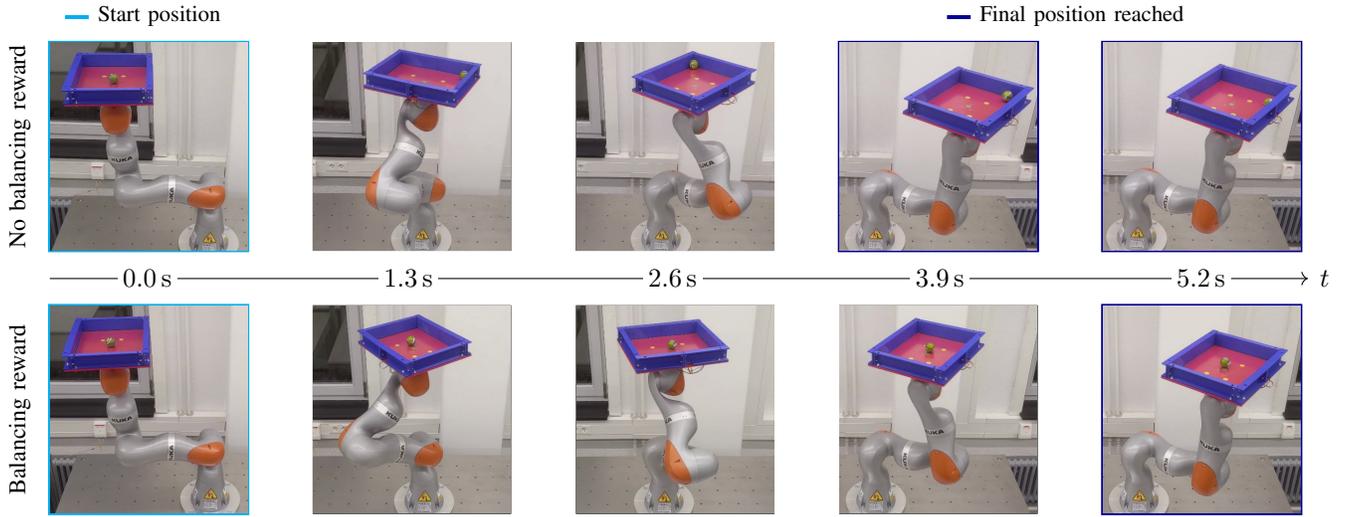}
	\caption{%
	The ball-on-plate task performed by a real KUKA iiwa. 
	When the balancing performance is rewarded (bottom), the reference path is traversed less quickly, but the ball is kept close to the center of the plate. 
	}
	\label{fig:real_robot_balancing}
	\vspace{-0.45cm}
\end{figure*} 
While our method additionally supports jerk limits and online adjustments of the reference path, the offline method TOPP-RA generates faster trajectories that track the reference path almost perfectly.  
A quantitative analysis is shown in TABLE \ref{table:toppra_timing}.  
Trajectories for the KUKA iiwa generated with TOPP-RA require around \SI{78}{\percent} of the time needed by our method. 
For the humanoid \mbox{ARMAR-4}, around \SI{65}{\percent} of the time is needed. 
In return, the reference paths need to be known in advance and it is not possible to consider additional objectives that require sensory feedback. 

\subsection{Path tracking with objectives based on sensory feedback}
\subsubsection{Ball-on-plate task}
The goal of the ball-on-plate task is to traverse a reference path while balancing a ball. 
During the motion execution, the position of the ball on the plate is provided as sensory feedback. 
TABLE \ref{table:ball_balancing} shows the training results with and without an additional reward component based on the balancing performance. 
Without the balancing reward, the ball falls off the plate in all evaluated episodes.
However, when the balancing performance is rewarded, the ball falls down in only \SI{0.3}{\percent} of the episodes.
In return, the reference path is traversed less quickly and less precisely.
Fig. \ref{fig:real_robot_balancing} visualizes the results based on an exemplary reference path traversed by a real KUKA iiwa. The robot is controlled using the networks trained in simulation and the ball position is provided by a resistive touch panel.
\begin{table}[b]
    \vspace{-0.1cm}
    \caption{%
    Comparison with the offline method \mbox{TOPP-RA.}
    }
    \vspace{-0.15cm}
    \makegapedcells
\begin{tabular*}{0.49\textwidth}{p{22.0mm}p{17.5mm}p{14.5mm}p{18.0mm}} 
    \toprule
& \hfil Duration [s] \hfil&  \hfil Relative \hfil &  \hfil Max. joint pos. \hfil \\
& \hfil (robot stopped) \hfil & \hfil duration [\%] \hfil & \hfil deviation [rad] \hfil \\
 \hline
KUKA iiwa & & & \\
\tabitem Random
    &  \hfil $3.36$ \hfil  &   \hfil $78.5$ \hfil   &  \hfil $0.00$ \hfil \\
\tabitem Target point 
    &  \hfil $3.79$ \hfil  &   \hfil $76.0$ \hfil   &  \hfil $0.00$ \hfil \\
\tabitem Ball balancing
    &  \hfil $1.91$ \hfil  &   \hfil $78.3$ \hfil   &  \hfil $0.00$ \hfil \\
\mbox{ARMAR-4 (fixed base)}& & & \\
\tabitem Random
    &  \hfil $3.31$ \hfil  &   \hfil $65.0$ \hfil   &  \hfil $0.00$ \hfil \\
\tabitem Target point 
    &  \hfil $3.68$ \hfil  &   \hfil $67.2$ \hfil   &  \hfil $0.00$ \hfil \\
    \bottomrule
    \end{tabular*}
    \vspace{-0.0cm}
    
\label{table:toppra_timing}
\end{table}

\begin{table}[t]
    \vspace{0.05cm}
    \caption{%
    Training results for the ball-on-plate task. 
    }
    \vspace{-0.15cm}
    \makegapedcells
\begin{tabular*}{0.49\textwidth}{@{}p{19.2mm}p{14.95mm}p{11.0mm}p{10.5mm}p{14.5mm}} 
    \toprule
\hspace{0.05cm}Ball balancing & \multicolumn{1}{c}{Duration [s]} & \multicolumn{1}{c}{Balancing} & \multicolumn{2}{c}{Cart. pos. deviation [cm]} \\
\hspace{0.05cm}dataset  & \hfil (end of path) \hfil & \hfil error [\%] \hfil & \multicolumn{1}{c}{mean} & \multicolumn{1}{c}{max} \\
 \hline
\hspace{0.05cm}KUKA iiwa & & & & \\
\hspace{0.00cm}\tabitem No balancing \newline \hspace*{0.00cm}\phantom{\tabitem }reward
    &  \hfil $2.24$ \hfil  &   \hfil $100.0$ \hfil   &   \hfil $1.4$ \hfil &  \hfil $3.0$ \hfil\\
\hspace{0.00cm}\tabitem Balancing \newline \hspace*{0.00cm}\phantom{\tabitem }reward
    &  \hfil $2.99$ \hfil  &   \hfil $0.3$ \hfil   &   \hfil $2.3$ \hfil &  \hfil $4.9$ \hfil\\
    \bottomrule
    \end{tabular*}
    \vspace{-0.56cm}
    
\label{table:ball_balancing}
\end{table}
\begin{table}[b]
    \vspace{-0.3cm}
    \caption{%
    Maintaining balance with ARMAR-4.
    }
    \vspace{-0.15cm}
    \makegapedcells
\begin{tabular*}{0.49\textwidth}{@{}p{19.2mm}p{14.95mm}p{11.0mm}p{10.5mm}p{14.5mm}} 
    \toprule
\hspace{0.05cm}Target point & \multicolumn{1}{c}{Duration [s]} & \multicolumn{1}{c}{Balancing} & \multicolumn{2}{c}{Cart. pos. deviation [cm]} \\
\hspace{0.05cm}dataset  & \hfil (end of path) \hfil & \hfil error [\%] \hfil & \multicolumn{1}{c}{mean} & \multicolumn{1}{c}{max} \\
 \hline
\hspace{0.05cm}ARMAR-4 with \newline \hspace*{0.05cm}fixed legs& & & & \\
\hspace{0.00cm}\tabitem No balancing \newline \hspace*{0.00cm}\phantom{\tabitem }reward
    &  \hfil $5.28$ \hfil  &   \hfil $26.1$ \hfil   &   \hfil $3.6$ \hfil &  \hfil $8.8$ \hfil\\
\hspace{0.00cm}\tabitem Balancing \newline \hspace*{0.00cm}\phantom{\tabitem }reward
    &  \hfil $5.63$ \hfil  &   \hfil $5.3$ \hfil   &   \hfil $4.1$ \hfil &  \hfil $9.6$ \hfil\\
\hspace{0.05cm}ARMAR-4 with \newline \hspace*{0.05cm}controlled legs& & & & \\
\hspace{0.00cm}\tabitem Balancing \newline \hspace*{0.00cm}\phantom{\tabitem }reward
    &  \hfil $5.59$ \hfil  &   \hfil $0.8$ \hfil   &   \hfil $4.2$ \hfil &  \hfil $9.8$ \hfil\\
    \bottomrule
    \end{tabular*}
    \vspace{-0.0cm}
    
\label{table:armar_balancing}
\end{table}

\subsubsection{Maintaining balance with ARMAR-4} 
The additional objective of this task is to prevent a bipedal robot from falling over.
For that purpose, the pose of the robot's pelvis is provided as sensory feedback.  
TABLE \ref{table:armar_balancing} shows the training results for three different experimental configurations.
In the first two experiments, the legs of the robot are fixed in an outstretched position.
Without an additional balancing term in the reward function, the robot falls over in \SI{26}{\percent} of the episodes. 
With an additional reward for standing upright, the robot loses balance in  approximately \SI{5}{\percent} of the episodes. 
In the third experiment, the 12 joints of the legs are also controlled by the neural network. %
Thus, the network can use the legs to stabilize the motions of the upper body. As a result, the robot falls over in less than \SI{1}{\percent} of the episodes. 
\begin{figure}[t]
	\hspace{-0.6cm}
	\resizebox{0.53\textwidth}{!}{
    \input{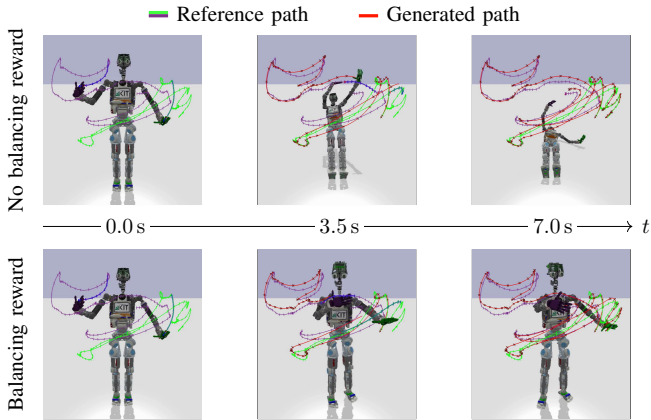}
    } %
	\caption{%
    Top: Without a balancing reward, the robot falls over. 
    Bottom: With a reward for standing upright, the legs of the robot are used to keep the robot in balance.
    }
	\label{fig:armar_balancing}
	\vspace{-0.45cm}
\end{figure}
An exemplary episode of the first and the third experiment is shown in Fig. \ref{fig:armar_balancing} and also in the accompanying video\href{\linkToVideo}{$^3$}. 

\subsection{Real-time capability}
As shown in Fig. \ref{fig:real_robot_balancing} and in the accompanying video\href{\linkToVideo}{$^3$},
we successfully applied our method to a real KUKA iiwa robot using networks trained in simulation.
The transfer is performed by sending the trajectory setpoints specified by each action to a real trajectory controller instead of a virtual one simulated by PyBullet. %
To analyze the computational requirements of our method, we calculate 1200 episodes for each of the three robots shown in this paper using an Intel i7-8700K CPU. 
We then calculate the quotient of the computation time and the trajectory duration and provide the highest value of all episodes in TABLE \ref{table:computation_times}. 
The results show that the computation time is significantly smaller than the trajectory duration, making our method well-suited for real-time trajectory generation.

\begin{table}[h]
    \vspace{-0.15cm}
    \caption{Evaluation of the computational effort.}
    \vspace{-0.15cm}
    \makegapedcells
\begin{tabular*}{0.49\textwidth}{@{}p{24.5mm}p{13.75mm}p{23.5mm}p{11mm}} 
    \toprule
\hspace{0.02cm}  &  KUKA iiwa &  ARMAR-4 with legs  & \multicolumn{1}{c}{ARMAR-6} \\
    \hline
\hspace{0.05cm}\scalebox{1.4}{$\frac{\text{Computation time}}{\text{Trajectory duration}}$}
    & \hfil \SI{7.50}{\percent}  & \hfil \SI{34.97}{\percent}     &  \hfill \SI{10.59}{\percent}   \\
    \bottomrule
    \end{tabular*}
    \vspace{-0.5cm}
\label{table:computation_times}
\end{table}
\section{Conclusion and future work}
This paper presented a learning-based approach to follow reference paths that can be changed during motion execution. 
Trajectories are generated by a neural network trained to maximize the traversing speed while minimizing the deviation from the reference path.
Additional task-specific objectives can be considered by including sensory feedback into the state.
The mapping of network actions to joint accelerations ensures that no kinematic joint limits are violated. 
We evaluated our method with and without additional objectives on robotic systems with up to 30 degrees of freedom 
showing that well-performing trajectories can be learned for reference paths with different path characteristics. %
We also demonstrated successful \mbox{sim-2-real} transfer for a ball-on-plate task performed by an industrial robot. %

In future work, we would like to investigate ways to additionally control the traversing speed during motion execution.

\section*{ACKNOWLEDGMENT}

This research was supported by the German Federal Ministry of Education and Research (BMBF) and the Indo-German Science \& Technology Centre (IGSTC) as part of the project TransLearn (01DQ19007A). We would like to thank Tamim Asfour for his valuable feedback and advice.

\bibliographystyle{IEEEtran}
\bibliography{root}

\end{document}